%% file: tst_arxiv_main.tex
\documentclass{article} 
\usepackage{iclr2026_conference,times}

\input{math_commands.tex}

\usepackage{hyperref}
\usepackage{url}

\usepackage[utf8]{inputenc} 
\usepackage[T1]{fontenc}    
\usepackage{hyperref}       
\usepackage{url}            
\usepackage{booktabs}       
\usepackage{amsfonts}       
\usepackage{nicefrac}       
\usepackage{microtype}      
\usepackage{xcolor}         

\usepackage{titletoc} 
\usepackage{booktabs}
\usepackage{graphicx}
\usepackage{multirow}
\usepackage{pifont}
\usepackage{array}
\usepackage{wrapfig,lipsum,booktabs}
\usepackage{adjustbox}
\usepackage{hyperref}
\usepackage{xspace}
\usepackage{wrapfig}
\usepackage{subcaption}
\usepackage{caption}
\usepackage{tabularx}

\newcommand{\supmat}{{\color{blue}Appendix}\xspace}
\newcommand{\appref}{{\color{blue}App.}\xspace}

\newcommand{\method}{\texttt{TST}\xspace}
\newcommand{\tsp}{\texttt{TST}\xspace}
\newcommand{\tsprgb}{\texttt{TST-MAE}\xspace}
\newcommand{\tspmm}{\texttt{TST-MM}\xspace}
\newcommand{\tspdino}{\texttt{TST-DINO}\xspace}
\newcommand{\cmark}{\ding{51}}%
\newcommand{\xmark}{\ding{55}}%
\newcommand{\weburl}{\textcolor{cyan}{\url{https://tst-vision.epfl.ch}}}

\definecolor{customgreen}{RGB}{192, 206, 182}
\definecolor{customblue}{RGB}{198, 231, 255}

\title{Multimodality as Supervision: \\ Self-Supervised Specialization to the Test Environment via Multimodality}

\author{
Kunal Pratap Singh$^{*}$ \quad Ali Garjani$^{*}$ \quad Rishubh Singh \quad Muhammad Uzair Khattak \quad Efe Tarhan \vspace{0.3em} \\
\textbf{Jason Toskov} \quad \textbf{Andrei Atanov} \quad \textbf{Oğuzhan Fatih Kar\textsuperscript{$\dagger$}} \quad \textbf{Amir Zamir} \vspace{0.3em} \\
Swiss Federal Institute of Technology Lausanne (EPFL) \quad 
\\ \\
\weburl
}

\iclrfinalcopy 
\begin{document}

\maketitle

\begin{figure}[h!]
\centering
\vspace{-1em}
\includegraphics[width=\columnwidth]{figures/teaser_fig_cam_ready_iclr.pdf}
\caption{
\textit{Left:} \textbf{Multimodality as Supervision.} The data sensed by an agent in an environment is often multimodal and, besides RGB images, can contain various modalities, such as depth, motion sensing, surface normals, tactile, etc. 
This enables \textit{Cross-Modal learning}, i.e., predicting the response of one sensor from another, as a method for \textit{self-supervised} pre-training. 
Many commonly used devices such as household robots, smartphones, and AR/VR headsets are already equipped with many sensors, making cross-modal learning readily applicable.  
\\
\textit{Right:} \textbf{Test-Space Training.}
We study cross-modal learning in a controlled setup where an agent moves through the physical world and collects unlabeled multimodal data, which we use to build a self-supervised representation.
A particularly interesting instantiation arises when we restrict the agent's world to a single physical space, the test space. We perform both self-supervised pre-training and evaluation within it. This yields a representation specialized to the test space, and reflects real-world applications in which an agent repeatedly operates within the same physical space (e.g., household robotics). 
We call this \textit{Test-Space Training} (TST). Generalist pre-training instead relies on large-scale external data, such as images from the internet or observations from many other spaces. 
\textit{Test-Space Training} (TST), with cross-modal learning, outperforms various generalist pre-training baselines~\citep{ oquab2023dinov2, radford2021clip}.
}
\label{fig:pull}
\vspace{-0.5em}
\end{figure}

\begin{abstract}
\makeatletter
\def\@makefnmark{}   
\footnotetext{\textsuperscript{$\dagger$} Work done while at EPFL. Now at Apple.} 
\makeatother

\textbf{Cross-modal learning}, i.e., learning to predict one modality from another, leverages multimodality as a source of self-supervision. Many practical applications, such as deploying a household robot, involve devices equipped with a rich set of sensors that can collect multimodal data in their deployment environment. This presents an opportunity to learn representations from this multimodal data through cross-modal learning, which is currently underutilized. 
Prior work has studied cross-modal learning on pre-collected internet-based datasets. However, these datasets provide only a fixed set of observations and no access to the underlying physical environments from which additional viewpoints or modalities could be acquired.

We study cross-modal learning in a setup where a multisensory agent moves through the physical world and collects multimodal data. This gives us control over multimodal data collection, including the agent's operating space, viewpoints, amount of data, and available modalities. 
Within this setup, a particularly interesting instantiation arises when we restrict the agent's world to a single physical space. We call this the test space, and perform both self-supervised pre-training and evaluation within it. This results in a \textbf{specialization} setup in which the representation is developed for a specific test space. It also reflects practical applications in which an agent repeatedly operates in a fixed physical space (e.g., household robotics). We call this Test-Space Training (\tsp).
We find that cross-modal learning on rich multimodal test-space data enables \tsp to achieve state-of-the-art performance in that space, outperforming generalist baselines pre-trained on large-scale internet-based datasets~\citep{oquab2023dinov2,radford2021clip}. 
This reduces the need for external internet-scale pre-training datasets where the deployment environment is known.
We further study the trade-off between scaling external data and scaling modalities and observe that in our setup, additional modalities can substitute for more external unimodal data. 
Finally, varying pre-training data enables a trade-off between a model's abilities to specialize to a test space and generalize to held-out spaces.

\end{abstract}

\section{Introduction.}
\label{sec:intro}
\textbf{Multimodality is a fundamental source of self-supervision}. Suppose you are a just-born agent with multiple sensory modalities, e.g., sight and sound, but no prior representation for what they mean. The system is in a clean-slate state. The two modalities provide you with two time-locked streams of information sensed from one underlying physical world. By merely \textbf{learning to predict one modality from the other}, you can start learning a representation for the modalities. This is referred to as ``\textbf{cross-modal learning}''(Fig~\ref{fig:pull}, left)  and requires no external supervision beyond the sensory modalities you came with. 

Applying cross-modal learning across a \textbf{constellation of diverse modalities} (sight, sound, haptic, motion, self-action, imitation, smell, etc.), through \textbf{well-designed pre-training objectives}, can lead to a rich representation without any external supervision, achieved solely through multimodality. This angle has been hypothesized by developmental psychologists as one of the key drivers behind efficient yet effective biological intelligence \cite{Smith2005SixLessons}. As many commonly used devices, e.g., iPhone~\citep{arkitscenes}, are already equipped with a rich array of sensors, this setup is readily applicable in the real world and currently underutilized.

In this work, we are interested in studying how far cross-modal learning can go in AI using today’s ML machinery. Prior work~\citep{bachmann2022multimae,4m21,Lu2023UnifiedIO2S} has studied this on pre-collected internet datasets~\citep{Changpinyo2021CC12M}, and shown impressive any-to-any multimodal capabilities and performance on downstream tasks. These datasets provide a fixed set of observations from the physical world. Although these observations can be filtered, we cannot access the underlying environments in which they were captured to acquire new viewpoints or modalities. 

We consider an alternative setup in our work: we spawn a multisensory agent in the physical world. 
This setup gives us control over multimodal data collection, including the environments, viewpoints, amount of data, and available modalities. 
The agent can move around its environment and collect multimodal data from its sensors~\footnote{We experiment with two setups. One is based solely on physical sensory modalities, and one expands to use features of pre-trained neural networks as additional modalities. They are described later.}. 
We then perform self-supervised pre-training through cross-modal learning on the multimodal data using the process described above. 
After pre-training, we evaluate whether the learned representation can be efficiently transferred to novel tasks (e.g., object segmentation from images). Within this setup, a particularly interesting instantiation is when we assume the entire world of the agent is a single physical environment, such as a house or a building. We refer to this environment as \textit{the test space}. In this setup, cross-modal pre-training and evaluation of the learned representation both happen in the test space. It results in a ``\textbf{specialization}'' setup   (Fig.~\ref{fig:pull}, right), where a performant model is expected to be developed for a specific test space.  As the setup is self-supervised, testing in the same space where pre-training took place is valid, as no task-specific annotations are used from the test space. We refer to this instantiation as \textit{Test-Space Training} (\tsp), and describe it in more detail in Sec.~\ref{sec:method}.

This instantiation setup has three key advantages. \textbf{1)} It provides a \textbf{controlled sandbox} to study the question of interest. It brings the world under our control, allowing us to systematically vary data size and diversity (Sec.~\ref{sec:role_of_data}), and perform modality scaling experiments (Sec.~\ref{sec:role_of_mm}).
\textbf{2)} As opposed to experimenting with internet-scale diverse data, this setup better resembles the developmental settings of babies and biological agents that inspired this formulation. They manage to develop rich intelligence from a relatively limited physical context around them~\cite{smith2011not,pereira2014bottom}, in contrast to needing data with the diversity of the entire world, as in internet-based, scaled datasets. \textbf{3)} It has \textbf{real-world deployment value}. The described setup represents the deployment scenario of many physical intelligent agents. A house robot often doesn’t leave the house where it is deployed; as long as it can perform effectively in that specific house, it doesn't matter whether it would readily generalize to all other houses in the world. The same applies to AR/VR goggles, specialized assistants, and other cases. In other words, \textbf{many cases of deployed AI have a limited context around them}, which is \textbf{at odds with the approach of wishing ``generalist” agents} that are \textbf{expected to solve all problems everywhere unnecessarily}. This also relates to a broader point on excessive generalism elaborated by biologists, such as Frans de Waal~\cite{de2016we}. 

A central design choice in this study is which modalities to include in the dictionary, as it will directly affect the richness of the learned representation~\citep{4m,Zamir2018Taskonomy}. We first experiment with a bare setup (Sec.~\ref{sec:howfar}) that includes only modalities easily acquired from hardware sensors, such as RGB images and depth. We find that even this bare set is effective, as it covers almost half of the spectrum between the two bounds of no-pretraining (scratch) and a fully-supervised model (Fig.~\ref{fig:tst_consolidated}) in the test space. However, this model falls behind state-of-the-art internet-based generalists such as DINOv2~\citep{oquab2023dinov2}. This suggests that the bare set of modalities, given the current model architecture and training objectives, is sub-optimal for replacing generalist models in deployment. 
We therefore extend the modality dictionary further (Sec.~\ref{sec:tst_vs_internet}) to include pseudo-modalities that are latent representations of powerful pre-trained neural networks, such as CLIP~\citep{radford2021clip} and Imagebind~\citep{Girdhar2023ImageBindOE}. Leveraging these modalities essentially corresponds to \textbf{distilling the pre-trained models and specializing them to the test space}. 
The multimodal representation trained with this extended set of modalities yields state-of-the-art results in the test space (Tab.~\ref{tab:max-mod-quant}), and, interestingly, outperforms all pseudolabel networks from which the latent representations (\appref~\ref{sec:pseudolabel_baseline}) were distilled.
This raises an interesting question: \textit{when the deployment space is known, can specialized pre-training in the test space be more effective than relying solely on off-the-shelf generalist models?}
It suggests a possibility that distilling powerful pseudolabels on just deployment space data can be sufficient for performant specialist models. Note that this setup does not correspond to a case in which no information or data beyond the test space was seen (see Sec.~\ref {sec:tst_vs_internet} for details). 
Future research on multimodal learning will hopefully lead to findings that close the gap to state-of-the-art and fully supervised models without the need for externally pre-trained models and data. 

Through various analyses and ablations, we contribute the following key findings:

\noindent$\bullet$ \textbf{Multimodality as (Self-)Supervision is effective}, specifically under a test-space specialization setting. Our experiments show that this approach is promising, and when augmented with modalities based on pre-trained models, it leads to state-of-the-art results in the test space. (Tab.~\ref{tab:max-mod-quant})

\noindent$\bullet$ \textbf{Scaling modalities in the test space versus scaling external unimodal data.} We find that scaling the number of modalities in the test space is more effective (Fig.~\ref{fig:mod-vs-data}) than scaling unimodal data from an external source.

\noindent$\bullet$ \textbf{Specialization-Generalization Tradeoff.} We explore how, with the same number of samples, the source of pre-training data can enable a model to exhibit varying abilities, from being specialized to a test space, to generalizing over a set of held-out spaces (Fig.~\ref{fig:spec-gen-trade-off}). 



\section{Related work}
\label{sec:rel_work}
\noindent\textbf{Multimodal learning} aims to build models that can relate information from different modalities of underlying reality~\citep{Baltruaitis2017MultimodalML}. This can involve training separate encoders or a unified model on various modalities, like image, video, 3D, text, audio, etc.~\citep{arandjelovic2017look, lu2019vilbert, jaegle2021perceiverIO, radford2021clip, Girdhar2022Omnivore, Lu2022UnifiedIOAU, Lu2023UnifiedIO2S,Girdhar2023ImageBindOE}. 
MultiMAE~\citep{bachmann2022multimae} uses multimodal masked modeling to learn cross-modal predictive coding across multiple modalities. 
4M~\citep{4m,4m21} extends this idea further to train a multimodal foundation model across tens of modalities by leveraging discrete tokenizers. Similarly, the Unified-IO line of work~\citep{Lu2022UnifiedIOAU,Lu2023UnifiedIO2S} leverages discrete tokenizers to build autoregressive multimodal models that can generate images, text, audio, and actions.
These approaches build on pre-collected, internet-based datasets~\citep{Changpinyo2021CC12M,kakaobrain2022coyo700m, Schuhmann2022LAION5B} with fixed set of observations and modalities. 
In contrast, we consider a setting in which an agent has access to the underlying physical world and can actively collect multimodal observations. 
This physical access allows multimodal signals available through the agent's onboard sensors to directly serve as sources of self-supervision for learning visual representations, while also giving us control over the space, viewpoints, amount of data, and available modalities. Together, these properties provide a controlled setting to study multimodality as a source of self-supervision. 
Within this setting, we study a specialization setup in which we use cross-modal learning on multimodal observations collected by the agent to learn a representation specialized to its deployment space.

\textbf{Self-supervised learning~(SSL)} has been effective in learning visual~\citep{ chen2020simple, He2021MaskedAA,Bao2021BEiT, oquab2023dinov2} and natural language~\citep{Devlin2019BERT, Brown2020GPT3, openai2023gpt4} representations. 
In vision, one line of work uses masked image modeling as a scalable approach to pre-train self-supervised models. It masks an input image, and attempts to reconstruct it in the form of pixels~\citep{He2021MaskedAA,chen_generative_2020, Dosovitskiy2020vit, ElNouby2021AreLD}, tokens~\citep{Bao2021BEiT} or features~\citep{zhou2021ibot,baevski2022data2vec}. On the other hand, approaches like SimCLR~\citep{chen2020big} and DINOv2~\citep{oquab2023dinov2} use contrastive learning~\citep{caron2021emerging,chen2020big,He_2020_CVPR,Chen2020ExploringSS} and knowledge distillation ~\citep{bucilua2006model,hinton2015distilling} respectively, to pre-train representations.  
The self-supervised approaches discussed above derive supervision from the structure of visual data itself, whereas we study cross-modal learning as a source of self-supervision, where we learn by predicting one modality from another. Many commonly used devices such as smartphones are already equipped with a rich array of sensors, making cross-modal learning naturally applicable in the real world.  

\noindent\textbf{Pre-training data} has been a critical component in the recent advances in self-supervised~\citep{oquab2023dinov2,caron2021emerging} and multimodal foundation models~\citep{4m,openai2023gpt4,Ramesh2021DALLE1}. Prior work~\citep{gadre2024datacomp,li2024datacomplm,Fang2023DataFN} has studied the role of pre-training data curation criteria in large-scale image-text models~\citep{radford2021clip}. Similarly, \citep{el2021large} investigates the role of external pre-training data for downstream performance on various target transfer tasks~\citep{Lin2014MSCOCO,Zhou2017ADE20K}. They show that pre-training via masked modeling directly on just the target task images performs on par with pre-training on a large-scale external dataset such as Imagenet~\citep{Russakovsky2014ImageNet}. 
Recent work on specialist pre-training in language models~\citep{Baek2026TheFF} shows that incorporating a small amount of data from the downstream deployment domain during pre-training can improve performance in that domain.
Our work studies a different form of specialization, where the target is a physical space rather than a downstream task or dataset. This setting reflects practical deployment scenarios in which devices such as smartphones and AR/VR headsets repeatedly operate in, and collect data from, a particular physical environment.


\noindent\textbf{Test-time adaptation} adapts a model to distribution shifts at test-time~\citep{Xiao2024BeyondMA}. One prominent approach in the community is test-time training (TTT)~\citep{sun19ttt, Wang2021TentFT,liu2021tttplus,Gandelsman2022TestTimeTW,boudiaf2022parameter,gao2023back}, which optimizes a self-supervised objective at test-time to adapt the model. 
In contrast, \tsp uses data from a given test space during pre-training to learn a representation specialized to that space, rather than adapting the model during downstream evaluation. TTT is complementary to TST, and we show in \appref\ref{appendix:ttt-with-tsp} that applying it on top of \tsp can further improve performance.  

We present additional related works in \appref\ref{appendix:additional_related_work} on domain adaptation, embodied active learning, and semi-supervised learning.

\section{Method}
\label{sec:method}
\begin{figure*}[t]
\centering
\includegraphics[width=\textwidth]{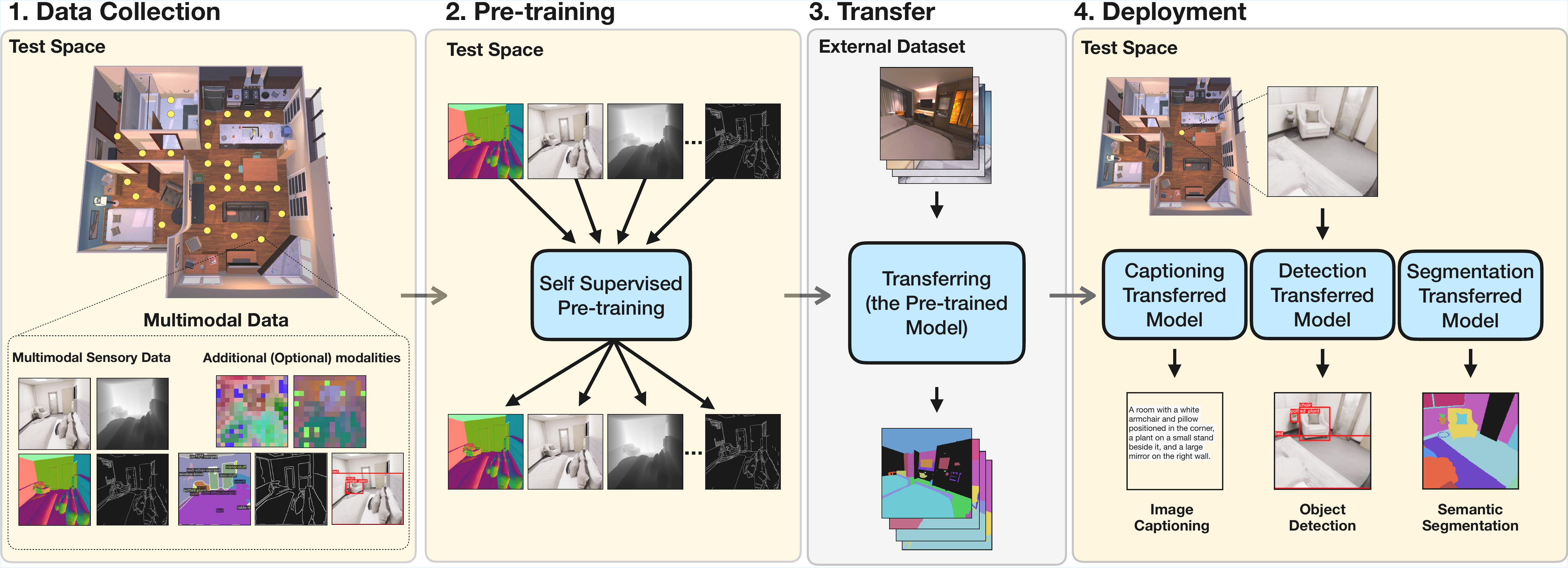}
\caption{\textbf{Test-Space Training.} 
\textbf{1)} First, we collect (multimodal) data from the test space~(Sec.~\ref{sec:data}). 
\textbf{2)} We then use this data for self-supervised multimodal pre-training~\citep{4m,oquab2023dinov2}~(Sec.~\ref{sec:crossmodal}).
\textbf{3)} After pre-training, the model is fine-tuned on a small external transfer dataset to solve a desired downstream task, e.g., semantic segmentation~(Sec.~\ref{sec:evaluation}). 
\textbf{4)} This model is subsequently deployed and evaluated in the same test space where it was pre-trained~(Sec.~\ref{sec:result}).
}
\vspace{-1em}
\label{fig:method}
\end{figure*}

We study cross-modal learning as a source of self-supervision for representation learning. To do so, we propose a controlled sandbox, in which an agent moves around and collects multimodal data in the physical world. We then learn self-supervised representations through cross-modal learning on this data. 
We focus on a particularly interesting instantiation, Test-Space Training (\tsp; Sec.~\ref{sec:tst_method}), which assumes that the agent's entire world consists of a single physical space, e.g., a house or a building.
In \tsp, self-supervised pre-training and downstream evaluation both take place in this test space.
This is a valid assumption, as the pre-training is self-supervised and does not assume any task annotations from the test space. \tsp therefore yields a \textbf{specialization} setup in which the objective is to learn a model that is performant in a specific test space. This setup mirrors real-world deployment scenarios for various user devices such as household robots and AR/VR goggles, where the primary objective is performance in a particular user environment. 

\subsection{Test-Space Training}
\label{sec:tst_method}
We present an overview of the Test-Space Training (\tsp) setup in Fig.~\ref{fig:method}.
It starts by collecting unsupervised, multimodal pre-training data in the test space. 
Concretely, we assume access to the sensory data sampling function in the test space, denoted as $x \sim p_{\mathrm{space}}(x)$, and use it to collect a pre-training dataset $D_{PT} = \{x_i\}$~(Sec.~\ref{sec:data}). Besides RGB images, we also leverage other sensors available on the device, e.g., depth and surface normals. 
In real-world deployment, this set can be expanded significantly to other common sensors, such as IMU, microphone, radar, and haptics. We use this data to pre-train a self-supervised model $f: X \to h$ that maps RGB images into representations~(Sec.~\ref {sec:crossmodal}). We evaluate this model on downstream tasks in the same test space via transfer learning (Sec.~\ref{sec:evaluation}).  

\subsubsection{Multimodal Data collection}
\label{sec:data}
To construct $D_{PT}$, we collect multimodal observations from the test space. This can correspond to capturing from various vantage points~\citep{Eftekhar2021Omnidata}, or a video walkthrough~\citep{arkitscenes,yeshwanthliu2023scannetpp} to cover the test space. 
Additionally, we can also process this data to create more optional modalities as illustrated in Fig.~\ref{fig:method}. As we later show in Sec.~\ref{sec:role_of_mm} and Fig.~\ref{fig:mod-vs-data}, scaling this rich set of modalities in the multimodal, test-space data is more effective than scaling to diverse unimodal data from external sources.
It is also worth noting that such a dataset of potentially repetitive images from the same space is related to findings in developmental psychology research suggesting that infants observe highly redundant visual data~\citep{jayaraman2015faces,slone2019self}.
We defer more implementation details for our data collection to Sec.~\ref{sec:setup}. This stage results in a multimodal dataset, which we use for cross-modal learning.

\subsubsection{Cross-modal Pre-training}
\label{sec:crossmodal}
We pre-train an encoder-decoder transformer~\citep{Vaswani2017AttentionIA} model on $D_{PT}$ collected in the test space using cross-modal learning. In this work, we implement cross-modal learning using multimodal masked modelling~\citep{4m,4m21} and refer to the resulting variant of \tsp as \tspmm. We provide more implementation details on the model naming notation, architecture, and objective in Sec.~\ref{sec:setup}. After pre-training, we retain the encoder $f$, which transforms RGB input to representations used for downstream transfer. 


To study various aspects of cross-modal learning, we begin with the choice of modalities in \tspmm. This is a critical choice, as it directly influences the quality of representation learned by the model. We explore a minimal setup with only hardware-based modalities\footnote{we describe the setup in more detail in Sec.~\ref{sec:setup}} in Sec.~\ref{sec:howfar}, and scale them further to leverage internet-based pseudolabel networks as additional modalities in Sec.~\ref{sec:tst_vs_internet}. We find that \tspmm, with the scaled set of modalities, achieves state-of-the-art performance in the test space (Tab.~\ref{tab:max-mod-quant}). We also show various other analyses on the role of multimodality (Sec.~\ref{sec:role_of_mm}) and pre-training data source (Sec.~\ref{sec:role_of_data}) in specialization.

It is worth noting that although the objective of this work is to study cross-modal learning, \tsp can also support other self-supervised objectives. We explore unimodal learning (RGB-only) objectives~\citep{He2021MaskedAA,oquab2023dinov2}, and find that multimodal learning is the most performant in this setup, consistently achieving superior performance (\appref\ref{appendix:benchmark_ssl_obj}).

\subsubsection{Evaluation via Transfer Learning}
\label{sec:evaluation}
We evaluate the effectiveness of the pre-trained model $f$ using transfer learning. We add a task-specific head $g$ and fine-tune the resulting model $g \circ f$ on various downstream tasks, following standard practice in self-supervised learning~\citep{4m,He2021MaskedAA}. For this, we consider a small transfer dataset $D_t$ with task-specific annotations, collected in an external space\footnote{This is a practical choice as it can be infeasible for a user to collect data in their own space.}, disconnected from the test space. Importantly, \textit{we do not have access to any task-specific annotations from the test space itself}, i.e., $D_t$ and $D_{PT}$ are sampled from different distributions. We benchmark against several off-the-shelf vision models~\citep{radford2021clip,oquab2023dinov2,4m21}, by finetuning them on the transfer data, $D_t$, as discussed in Sec.~\ref{sec:tst_vs_internet}.

\section{Experiments}
\label{sec:result}
We present the results as follows. Sec.~\ref{sec:setup} describes our experimental setup and baselines. 
Sec.~\ref{sec:howfar} starts with a minimal setup, which only assumes access to hardware-based modalities. This setup is fairly performant, but leaves room for further improvement compared to internet-based generalists. 
Therefore, in Sec.~\ref{sec:tst_vs_internet}, we extend the modality dictionary by leveraging off-the-shelf pseudolabel networks, and find that this leads to state-of-the-art models for the test space (Tab.~\ref{tab:max-mod-quant}).
Next, we leverage our sandbox to draw insights on the role of multimodality (Sec.~\ref{sec:role_of_mm}) in building specialized models with \tsp. Lastly, Sec.~\ref{sec:role_of_data} explores the role of pre-training data from the test space in achieving specialization. 



\subsection{Experimental Setup}
\label{sec:setup}
\noindent\textbf{Datasets.} We experiment using three datasets:\\
\noindent \textit{1. Scannet++}~\citep{yeshwanthliu2023scannetpp} is a large-scale dataset of \emph{\textbf{real-world}} indoor spaces containing sub-millimeter resolution scans, paired with DSLR and iPhone RGB images. We use 8 scenes as the test space and use a mix of iPhone and DSLR images from these scenes for pre-training. \\
\noindent \textit{2. Replica}~\citep{straub2019replica} provides high-quality 3D reconstructions of \textit{real} indoor spaces. We use 5 scenes as the test space and use rendered RGB-D images for pre-training. \\
\noindent \textit{3. ProcTHOR}~\citep{procthor} includes procedurally generated house-like environments. We use 5 procedurally generated houses as the test space, unless specified otherwise. 

\noindent\textbf{Pre-training.} 
For training \tspmm models on the dataset $D_{PT}$ collected from a test space, we implement cross-modal learning via multimodal masked modelling~\citep{4m} and train an encoder-decoder transformer model as described in Sec.~\ref{sec:crossmodal}. Following 4M~\citep{4m}, we use modality-specific tokenizers to convert all modalities into discrete tokens. At each training step, we sample a subset of tokens across the available modalities as input, and another subset of the remaining tokens as outputs. The encoder processes the input tokens, and the decoder predicts the target tokens conditioned on the encoder outputs. We optimize a cross-entropy loss over the target-token predictions.  

We train models across two encoder sizes, ViT-S and ViT-B, which have 8 and 12 encoder layers, respectively. 
Additionally, we found that mixing RGB images from the transfer was beneficial in pre-training. Please see \appref~\ref{appendix:transfer-mix-in} for an ablation on this choice. Note that \textit{we do not use any task labels from the transfer set during pre-training, making this stage task-agnostic.} 
We train \tspmm models from scratch, except for the adaptation experiments ((Sec.~\ref{sec:tst_vs_internet}, Adaptation through \tsp)) where we initialize from 4M-21.

\noindent\textbf{Notations.} We refer to \tsp variants with different objectives as \tspmm for multimodal masked modeling, \tsprgb for unimodal masked modeling, \tspdino for DINOv2~\citep{oquab2023dinov2}. Unless specified otherwise, we refer to the multimodal version (\tspmm) as \tsp.

\noindent\textbf{Transfer and Evaluation.} For all datasets, we use an external set of scenes that are different from the test space to collect a small transfer set ($D_t$) with task-specific annotations. We evaluate the transferred models in the test space on semantic segmentation (Scannet++, Replica, ProcTHOR), object detection (Scannet++, ProcTHOR), and image captioning (ProcTHOR). 

\noindent\textbf{Modalities.} For Scannet++~\citep{yeshwanthliu2023scannetpp}, we use RGB images captured by DSLR and iPhone cameras. For Replica~\citep{straub2019replica} and ProcTHOR~\citep{procthor}, we render RGB and Depth from the test space using onboard sensors. We also include Surface normals and Canny edges, as they can be extracted from RGB and depth, respectively, using simple operations. We refer to these 4 modalities (RGB, Depth, Surface normals, and Canny Edges) as \emph{sensory}.
In Sec.~\ref{sec:tst_vs_internet}, we discuss how we further scale this modality set using pseudo-modalities using off-the-shelf networks.

\noindent\textbf{Baselines.}
We compare against several baselines: \\ 
$\bullet$\noindent \textit{ Scratch - no pre-training.} We compare against both unimodal scratch, which takes only RGB images as input, and multimodal scratch, which inputs all modalities available to \tspmm during transfer training and evaluation. The latter baseline indicates that the performance is not owed to merely having multiple modalities, but rather performing cross-modal pre-training. \\
$\bullet$\noindent \textit{ Large-scale generalist pre-training baselines.} We evaluate 4M-21~\citep{4m21}, DINOv2~\citep{oquab2023dinov2}, and CLIP~\citep{radford2021clip} as recent strong generalist (self-supervised) baselines, trained on large-scale datasets via unimodal and multimodal learning. 
\\
$\bullet$\noindent \textit{ Task specialist baselines.} We perform evaluations using Mask2Former~\citep{cheng2021mask2former}, ViTDet~\citep{li2022vitdet}, SAM~\citep{Kirillov2023SAM}, and LLaVA-1.5~\citep{liu2023visual} as established task-specific baselines for semantic segmentation, object detection, and image captioning. 

We fine-tune all baselines with the same transfer data $D_t$, and use only RGB images as input for transfer and evaluation (except for the multimodal scratch baseline). An extensive hyperparameter search was done for fair comparison.
We provide more details on our experimental setup, including pre-training and transfer training hyperparameters in \appref~\ref{appendix:implementation-details}.

\begin{figure}[t!]
    \centering
    \includegraphics[width=0.75\linewidth]{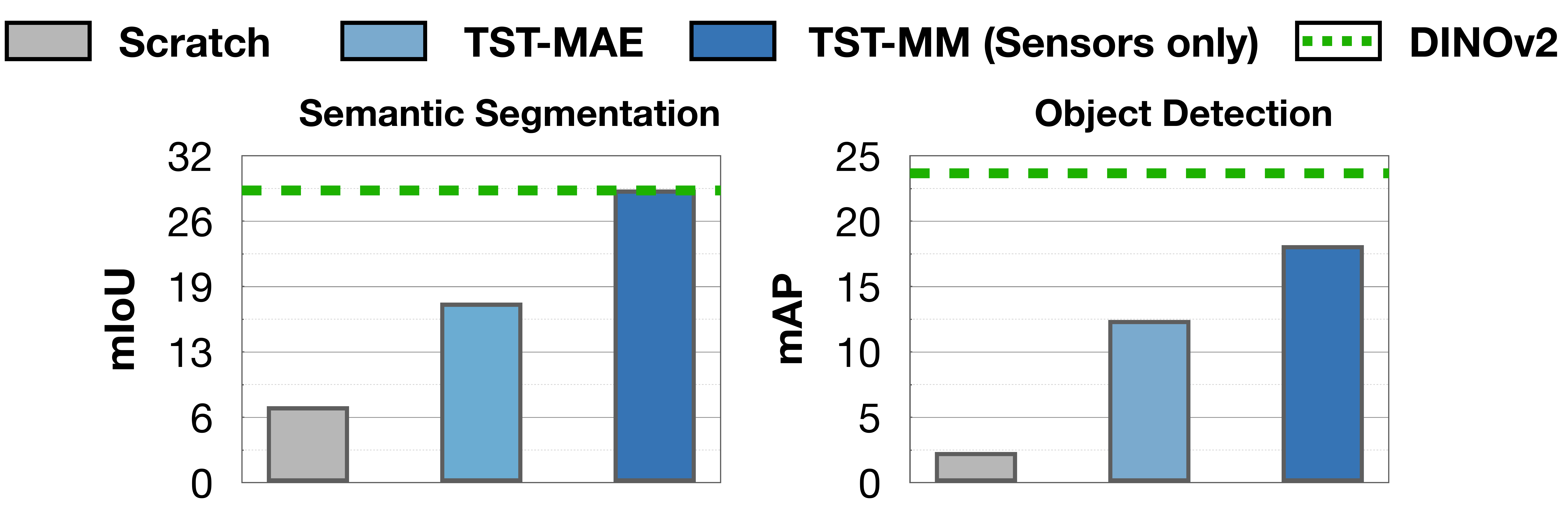}
    \caption{
        \textbf{How far can we go with no external access?}
        We compare \tspmm (Sensors) , trained using only data collected from the test space with onboard sensors, against DINOv2~\citep{oquab2023dinov2}, a strong off-the-shelf generalist pre-trained on 142M Internet images.
        We show segmentation and detection results on a test space from the Scannet++.
        \textit{We find that, with no external access, \tspmm with sensory modalities, and just multimodal data from the test space, is competitive with large-scale Internet-based pre-training.}
    }
    \label{fig:tst_sensors}
\end{figure}

\begin{figure}[t!]
    \centering
    \includegraphics[width=0.70\linewidth]{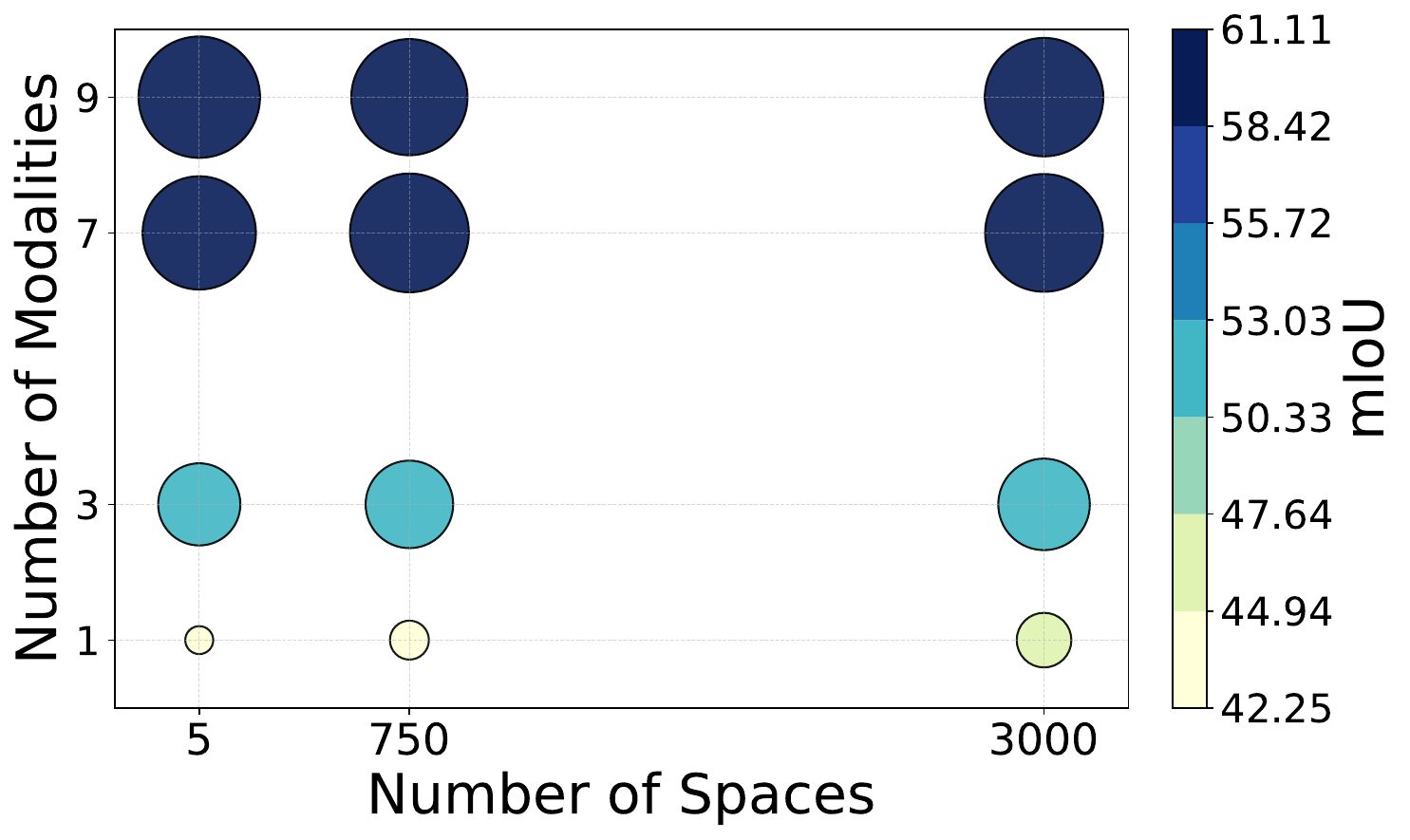}
    \caption{
        \textbf{Modality scaling vs data scaling.} 
        We study the trade-off between collecting unimodal pre-training data from more spaces versus scaling modalities in the test space (here, 5 houses). 
        The size of each circle is proportional to the mIoU performance on segmentation.
        We find that scaling the number of modalities within the test space yields better performance than scaling data by including external spaces. All models use the ViT-S backbone.
    }
    \label{fig:mod-vs-data}
\end{figure}

\subsection{How far can we go with no external access?}
\label{sec:howfar}
A central design choice in our setup is the choice of modalities to perform cross-modal learning over. It directly affects what representation is learned by the model.
We begin by exploring a setup that starts with a set of sensory modalities (as described in Sec.~\ref{sec:setup}) that can be acquired via hardware sensors.
This is a minimal setup that allows bootstrapping a vision representation for the test space, with \textit{no external access}. Fig.~\ref{fig:tst_sensors} shows that pre-training with just multimodal sensory data from a real-world test space can perform competitively with generalist models such as DINOv2~\citep{oquab2023dinov2}, pre-trained on large-scale Internet data. Additionally, although unimodal pre-training (\tsprgb) improves over learning from scratch, multimodality with \tspmm (Sensors only) performs the best.

This suggests that with cross-modal learning on just sensory modalities, \textit{we can build highly performant models for a test space, without any external access}, and achieve competitive performance with large-scale Internet-based pre-training~\citep{oquab2023dinov2}. However, the remaining performance gap on some downstream tasks (e.g., object detection) suggests that sensory modalities alone, under the current multimodal architecture and objective, do not consistently match internet-based generalists. 

\begin{table*}[t!]
\centering
\resizebox{\textwidth}{!}{%
\begin{tabular}{@{}llccccccc@{}}
\toprule
\multirow{3}{*}{}                                                                   & \multicolumn{1}{c}{\multirow{3}{*}{Method}} & \multicolumn{3}{c}{Semantic Segmentation}   & \multicolumn{2}{c}{Object Detection} & \multicolumn{2}{c}{Captioning} \\ \cmidrule(l){3-9} 
& \multicolumn{1}{c}{}                        & Scannet++ & ProcTHOR       & Replica        & Scannet++         & ProcTHOR         & \multicolumn{2}{c}{ProcTHOR}   \\
& \multicolumn{1}{c}{}                        & mIoU      & mIoU           & mIoU           & mAP               & mAP              & CIDEr          & SPICE         \\ \midrule
\multirow{2}{*}
{\begin{tabular}[c]{@{}l@{}} No Pre-training 
\end{tabular}} 
& Unimodal Scratch - no pre-training                   & 7.49      & 28.62          & 9.23           & 2.35              & 24.59            & 17.1           & 14.8    \\
& Multimodal Scratch - no pre-training                   & 7.82      & 26.29          & 10.03           & 3.76              & 19.19            & 11.0           & 10.5    \\ \midrule
\multirow{4}{*}
{\begin{tabular}[c]{@{}l@{}} Generalist \\ Pre-training 
\end{tabular}} 
& 4M (RGB-only) / MAE~\citep{He2021MaskedAA}                         & 13.74     & 46.29          & 18.18          & 18.31             & 37.17            & 30.4           & 19.1          \\
& 4M-21~\citep{4m21}                                       & 27.59     & 53.24          & 26.30          & 25.91             & 41.43            & 36.2           & 20.3          \\
& DINOv2~\citep{oquab2023dinov2}                                      & 30.60      & 54.50          & 26.72          & 23.67             & 40.28            & 14.7           & 13.5          \\
& CLIP~\citep{radford2021clip}                                        & 23.19     & 48.66          & 20.92          & 19.75             & 38.47            & 18.4           & 16.2          \\ \midrule
Task Specialist                                                                     & Task Specific Methods \footnotemark[2]                      & \underline{34.75}     & 56.72          & 28.51          & 23.59             & 44.10            & \textbf{40.60}           & \textbf{21.00}          \\ \midrule
\multirow{2}{*}{\begin{tabular}[c]{@{}l@{}} Specialist \\ Pre-training\end{tabular}} & \tspmm                                      & 34.49     & \textbf{60.85} & \underline{32.87} & \underline{31.54}    & \underline{49.38}   & 34.3           & 20.4          \\
& \tspmm (adapted)                            & \textbf{36.44}         & \underline{60.59}              & \textbf{34.53}              & \textbf{35.83}                 & \textbf{51.25}                & \underline{39.9}              & \underline{20.5}             \\ \bottomrule
\end{tabular}
}
\caption{
\textbf{Multimodal Test-Space Training (\tspmm) outperforms both strong generalists and task specialists across tasks.}
We evaluate semantic segmentation, object detection, and image captioning. All models use ViT-B backbones, except SAM~\citep{Kirillov2023SAM} (ViT-H). \tspmm (adapted) refers to fine-tuning 4M-21 on test-space data. On segmentation and detection, \tspmm consistently outperforms Internet-based generalists and matches or surpasses specialists. On captioning, \tspmm (from scratch), despite no text pre-training, matches 4M-21 trained on CC12M image-text pairs; \tspmm (adapted) surpasses 4M-21 and approaches LLaVA-1.5~\citep{liu2023visual}.}
\label{tab:max-mod-quant}
\end{table*}
\footnotetext[2]{Task-specific methods used for each result, in order: SAM~\citep{Kirillov2023SAM} (segmentation), ViTDet~\citep{li2022vitdet} (detection), and LLaVA~\citep{liu2023visual} (captioning)}

\subsection{Scaling Modalities in the Test Space}
\label{sec:tst_vs_internet}
Sec.~\ref{sec:howfar} discusses that although \tsp with cross-modal learning with only sensory modalities achieves competitive performance against internet-based generalists~\citep{oquab2023dinov2}, it is not performant enough to replace them yet.
Therefore, to enrich this representation further, we draw inspiration from recent progress in multimodal foundation models~\citep{bachmann2023scaling}, and scale up the set of modalities using off-the-shelf pseudolabel networks. Next, we compare \tsp with this scaled set of modalities (\tspmm), against Internet-based generalists and task specialist models. Lastly, we describe how \tsp can also enable adapting a pre-trained generalist to the test space.

\noindent\textbf{Additional Modalities.} We create new pseudo-modalities by pseudolabeling the collected RGB frames. We use neural network feature maps~\citep{radford2021clip,Girdhar2023ImageBindOE}, \textit{SAM edges}~\citep{Kirillov2023SAM}, \textit{bounding boxes} from ViTDet~\citep{li2022vitdet}, and \textit{semantic segmentation masks} from Mask2Former~\citep{cheng2021mask2former}. 
Note that \tspmm, which leverages these additional pseudo-modalities, no longer qualifies for \textit{no external access}, as it is akin to distilling the off-the-shelf pseudolabeling networks which were trained on large-scale external data. However, this distillation happens only on test-space data, therefore alleviating the need to directly access external data by leveraging off-the-model outputs as modalities. This is reminiscent of recent findings on attention transfer~\citep{Li2024OnTS}, which show that only distilling the attention patterns from a pre-trained teacher can achieve performance comparable to full fine-tuning of the pre-trained model. 

\noindent\textbf{\tsp vs. generalists.} Tab.~\ref{tab:max-mod-quant} shows quantitative results for \tspmm.
We compare against and outperform generalist models (MAE, DINOv2, 4M-21, and CLIP) trained on large-scale Internet datasets. \textit{This suggests that we can outperform generalist models by using cross-modal learning on multimodal data from the test space.} 
Figure~\ref{fig:maxmod-qual} shows qualitative improvements of \method over generalist Internet pre-training. Additionally, to ensure a fair comparison, we include the pseudolabeling networks used to generate pseudo-modalities as baselines. We show in \appref\ref{sec:pseudolabel_baseline} that \tspmm, trained \textit{from scratch}, outperforms them. 

\noindent\textbf{\tsp vs. task specialists.} We also show that \tspmm outperforms or is on par with off-the-shelf task specialist models on semantic segmentation~\citep{Kirillov2023SAM} and object detection~\citep{li2022vitdet}. For image captioning, despite not seeing any text data during pre-training, \tspmm performs on par with 4M-21~\citep{4m21}, which was pre-trained on large-scale image-text data~\citep{Changpinyo2021CC12M}, showing the effectiveness of the learned representation. 

\noindent\textbf{Adaptation through \tsp.} Tab.~\ref{tab:max-mod-quant} also presents results when we adapt an existing generalist model to the test space.
Unlike all \tspmm models discussed above that start from scratch, we start from a pre-trained 4M-21 model and fine-tune it on data from the test space, using multimodal masked modeling. The resulting model, \tspmm (adapted), significantly improves over 4M-21~\citep{4m21} in the test space. Therefore, \textit{\tsp can also serve as an adaptation mechanism for Internet-based models, making them more performant in the test space for downstream tasks.}

\subsection{Analyzing multimodality in TST}
\label{sec:role_of_mm}
We show that leveraging cross-modal learning, and specifically, multimodal masked modelling, is a key component in enabling performant models in the test space. 
In this section, we leverage the same sandbox setup (Sec.~\ref{sec:setup}) to perform controlled analysis on various aspects of multimodality in \tsp.

\noindent\textbf{Scaling data versus Scaling modalities?}
We study whether enriching \textit{test-space data with additional modalities} is more effective than scaling pre-training with \textit{large-scale unimodal external data}. 
We use ProcTHOR~\citep{procthor}, which allows us to procedurally generate house-like environments from the same underlying distribution as the test space. This allows us to scale external RGB data with similar spaces while minimizing domain shift relative to the test space.
Starting from unimodal pre-training in the test space, Fig.~\ref{fig:mod-vs-data} shows that scaling the number of modalities in the test space yields significantly better performance than increasing the amount of unimodal data from external sources. 
Additionally, we find that scaling unimodal external data improves over the unimodal baseline in the test space. However, for a higher number of modalities, we find that the performance is relatively stable when external data is added. All models here use a ViT-S backbone. 
This suggests: \textit{For building high-performing models in a specific test space, collecting data within that space using a richer set of modalities is often more effective than relying on large-scale, unimodal data collected from external sources.}

\begin{figure*}[t!]
\centering
\includegraphics[width=.9\textwidth]{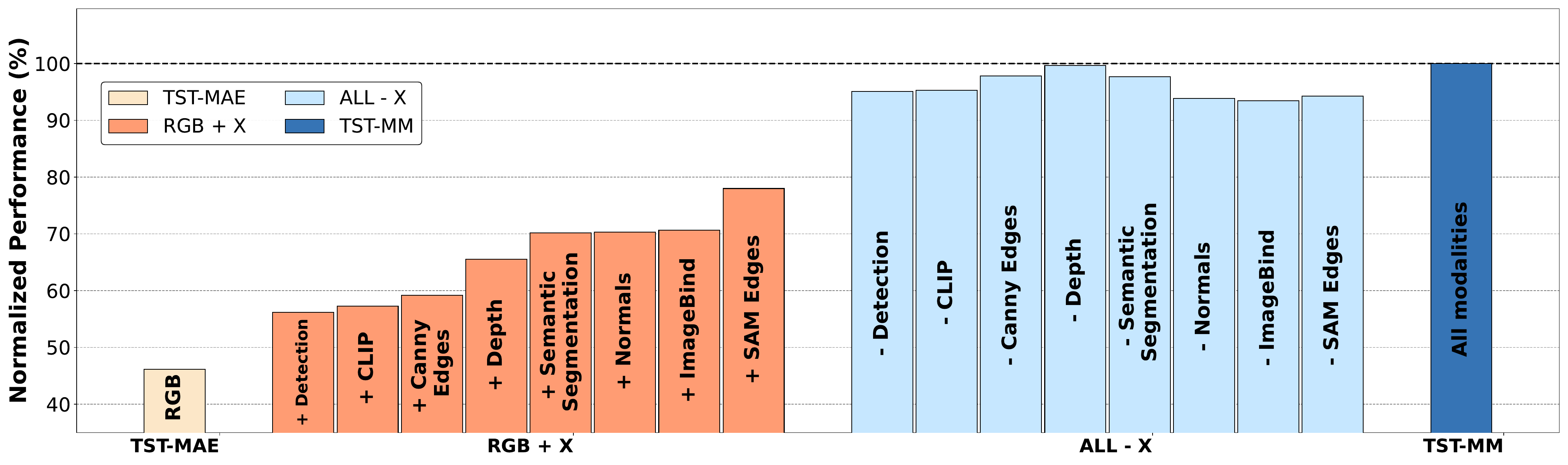}
\vspace{-0.9em}
\caption{
\textbf{Contribution of different modalities to \method performance.} 
We study the effect of each modality on \method by dropping one combination from \tspmm and adding one to \tsprgb~(RGB-only \method). We use the ViT-S backbone. We find that even though some modalities provide higher gains than others when added to the RGB-only \tsprgb, the performance of \tspmm stays relatively stable, agnostic to the choice of the dropped modality. \textit{This indicates that no single modality is responsible for \tspmm's performance, but rather their collective interplay, i.e., multimodality.}}
\label{fig:modality-engineering}
\vspace{-1em}
\end{figure*}



\noindent\textbf{Is the choice of modalities important for the effectiveness of the multimodal pre-training?}
We investigate whether all modalities contribute similarly to multimodal pre-training, or whether some contribute substantially more than others.
We present two ablations in Fig.~\ref{fig:modality-engineering}.
First, we examine all pairs of two modalities starting with RGB, i.e., all $\{\mathrm{RGB}, X\}$ combinations.
Adding any modality improves performance, with some showing greater benefits than others (e.g., \textit{SAM edges} increase performance by an absolute 7.8\%), but none matches the performance of using all modalities.
Second, we examine all sets of eight modalities by removing one modality from \tspmm, except RGB (which remains as the input during fine-tuning and evaluation).
We find low variance between different sets, indicating that \textit{no single modality is irreplaceable} and that other modalities can compensate for the absence of the missing ones.
For example, removing the \textit{SAM edges} modality reduces results by only 1.5\%, compared to its absolute 7.8\% improvement when added to RGB alone. \textit{Thus, within the \tspmm modality set, no single modality is irreplaceable.}


\noindent\textbf{How does the performance of \tspmm scale with modalities?}
Fig.~\ref{fig:modality_scaling} shows the performance of \tspmm as we increase the number of modalities.
Due to the combinatorial complexity of studying all possible combinations, we only sample all possible options for two (RGB+X) and eight \mbox{(All-X)} modalities, where X is the modality added or dropped. For other modality counts, we randomly sample 8  modality sets and report the average performance on the plot.
\textit{We find that the performance of \tspmm scales well with more modalities and shows relatively low variance across modality sets, which decreases as the number of modalities increases.}

\subsection{Measuring Specialization of TST}
\label{sec:role_of_data}
As discussed in Sec.~\ref {sec:setup}, our sandbox presents a \textbf{specialization} setup. 
A model is specialized to space A if it performs well in A, but is outperformed in another space B by a model specialized to B.

We define \textit{specialization} as the measure of how performant a model is on a downstream task, in a given test space, while forgoing other spaces. 
E.g., a model is specialized to space A if it performs well in A, but is outperformed in another space B by a model specialized to B. 
This is in contrast to \textit{generalization}, which we define as performance on held-out spaces that were not used in pre-training. 
As described in Sec.~\ref{sec:data}, \tsp collects data in a test space to pre-train a model for that space. 
In this section, we explore, given the same cross-modal learning objective, the role of this pre-training data from the test space itself. 


First, we measure specialization by cross-evaluating models pre-trained on three different test spaces, showing that space-specific pre-training performs the best.
Then, we explore whether we can substitute data from the test space with data from many (thousands of) similar spaces during pre-training, effectively asking: \textit{how many spaces is the test space worth?}
Third, we explore whether a single model can exhibit both specialization and generalization capabilities and show the \textit{specialization-generalization trade-off}. 

\begin{figure}[t!]
    \centering

    \begin{minipage}{0.48\linewidth}
        \centering
        \includegraphics[width=\linewidth]{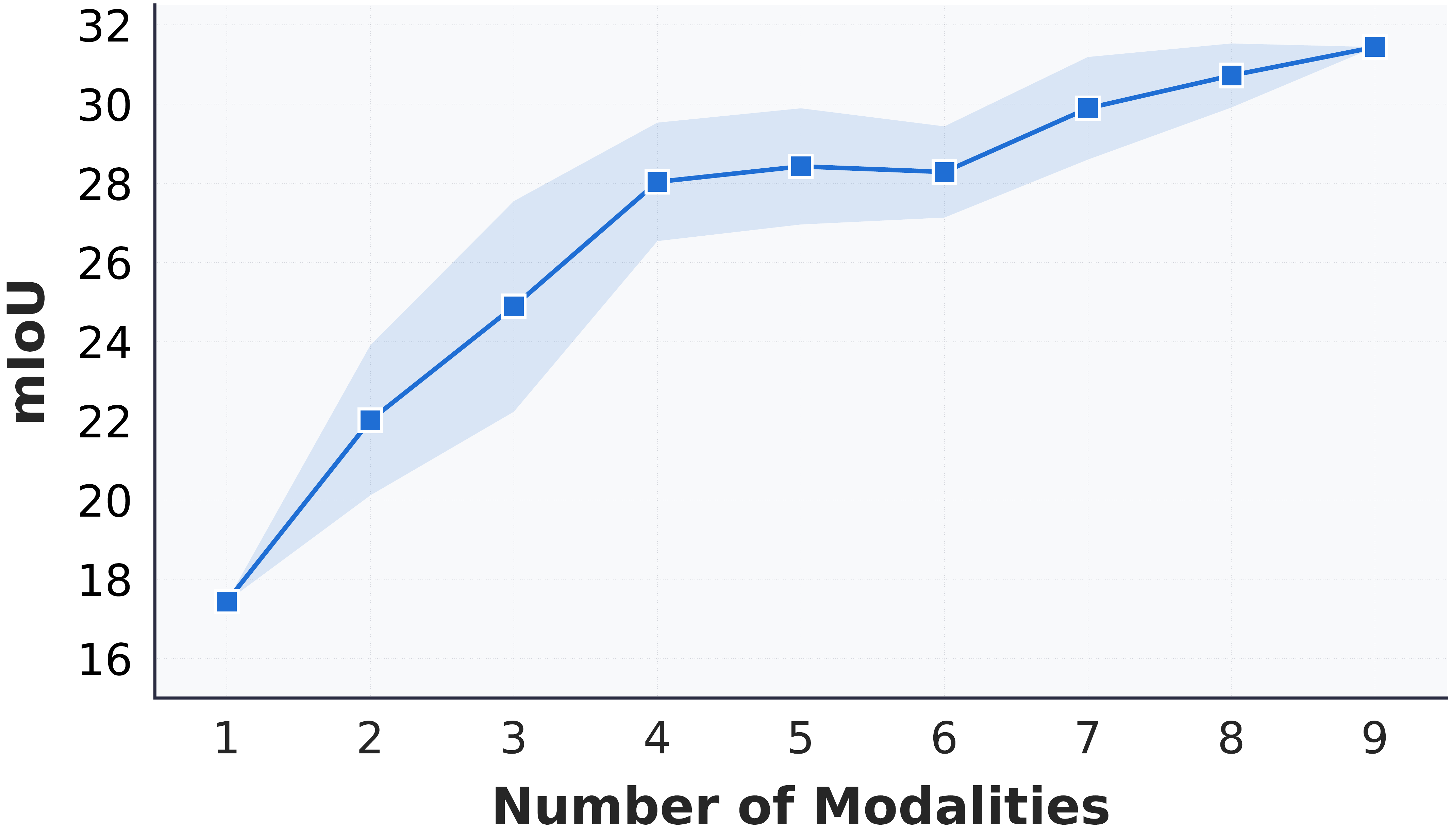}
        \caption{
            \textbf{Scaling the number of modalities for \tspmm.}
            We report the performance of \tspmm as we scale the number of modalities.
            We begin with only the RGB modality and incrementally add more.
            \textit{Increasing the number of modalities improves performance, while the variance in performance due to any specific modality decreases.}
        }
        \label{fig:modality_scaling}
    \end{minipage}
    \hfill
    \begin{minipage}{0.48\linewidth}
        \centering
        \includegraphics[width=0.9\linewidth]{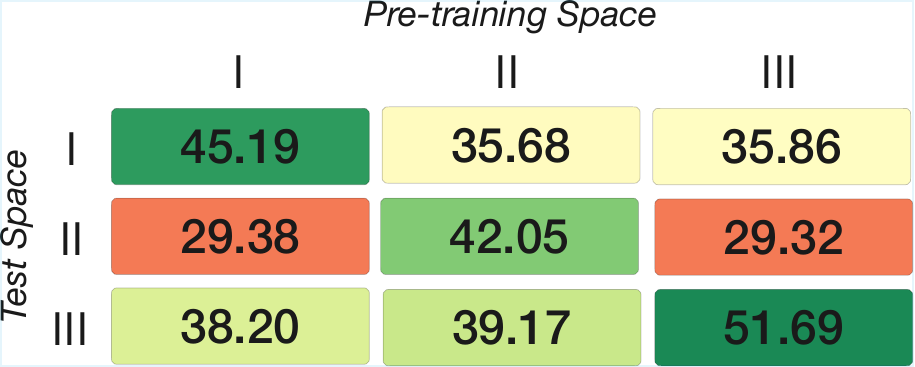}
        \caption{
            \textbf{Do we need the same test space for pre-training and evaluation?}
            We perform cross-space analysis by pre-training and evaluating performance on different spaces. 
            Each column and row represents a pre-training and test space. 
            \textit{Performance is best along the diagonal, where pre-training and evaluation are in the same space.}
        }
        \label{fig:1h-confusion}
    \end{minipage}

    \vspace{-0.8em}
\end{figure}

\begin{figure}[t!]
    \vspace{-0.5em}
    \centering
    \begin{minipage}{0.49\linewidth}
        \centering
        \includegraphics[width=0.9\linewidth]{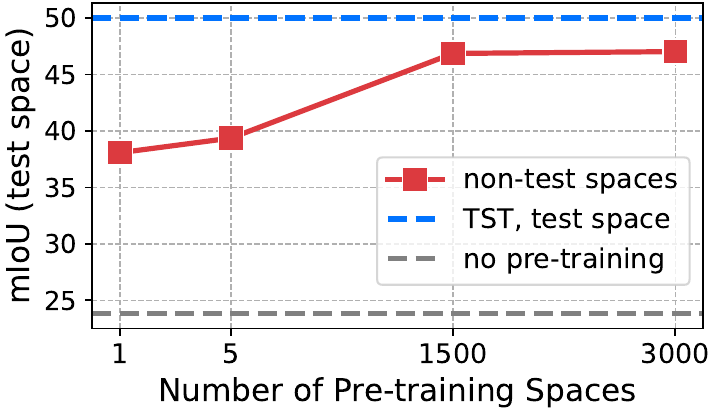}
        \caption{\textbf{How many spaces is one test space worth?}
        We study whether test-space data for pre-training can be substituted with data from similar but nonidentical spaces.
        We compare performance on the test space between \method and models pre-trained on an increasing number of IID houses. 
        The x-axis shows the four evaluated pre-training-space settings as categorical points and is not to scale.
        Both methods are trained with a ViT-S backbone. 
        \textit{We find that using as many as 3000 spaces cannot match pre-training in the exact test space}, underscoring the usefulness of test-space specialization with \method.}
        \label{fig:house-scaling}
    \end{minipage}
    \hfill
    \begin{minipage}{0.49\linewidth}
        \centering
        \includegraphics[width=0.9\linewidth]{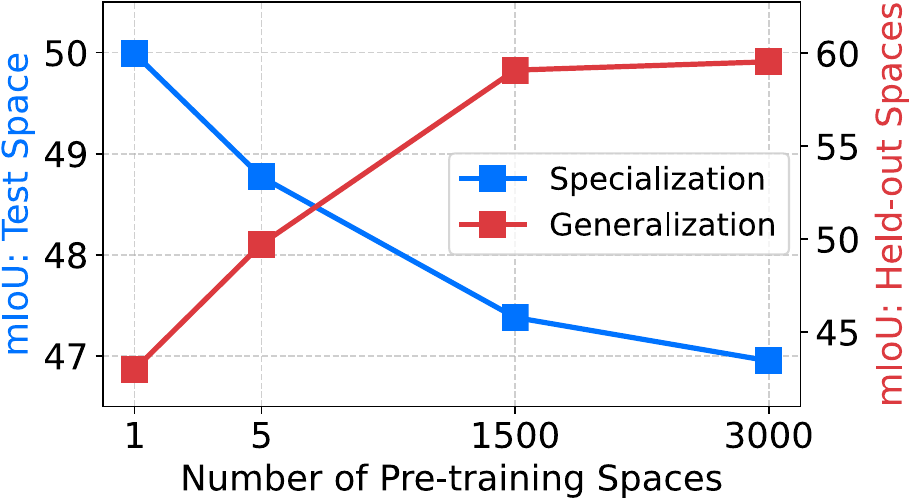}
        \caption{\textbf{\textcolor{blue}{Specialization}-\textcolor{red}{generalization} trade-off.}
        We pre-train ViT-S models on data collected from a growing number of spaces, starting with a single test space and adding data from other IID spaces.
        The blue curve shows performance in the test space (\textcolor{blue}{specialization}); the red curve shows performance on 100 held-out IID spaces (\textcolor{red}{generalization}).
        \textit{As we add more pre-training spaces, test-space performance decreases while held-out performance improves, revealing the \textbf{\textcolor{blue}{specialization}-\textcolor{red}{generalization} trade-off.}}}
        \label{fig:spec-gen-trade-off}
    \end{minipage}
    \vspace{-1.0em}
\end{figure}

\noindent\textbf{\method effectively specializes to a test space.}
Fig.~\ref{fig:1h-confusion} shows the performance of models pre-trained and evaluated on different test spaces.
We find that in all cases pre-training the model in the corresponding test space is the best, demonstrating that \tsp specializes to the test space.
We observe the same specialization trend across the other pre-training objectives (\appref~\ref{appendix:other-ssl}), suggesting that specialization to a test space is a more general phenomenon. However, across different objectives, we find cross-modal learning to be the best (\appref~\ref{appendix:benchmark_ssl_obj}). 

\noindent\textbf{How many spaces is a single test space worth?}
We show in Sec.~\ref{sec:tst_vs_internet} that \tsp outperforms internet-based generalists in the test space. 
However, general images from the Internet can be out-of-distribution (OOD) to the test space. 
In our cross-space analysis above, we remove this distribution shift by comparing against data from a similar (IID) space and find that \tsp is still the best. 
In this analysis, we take this a step further and compare against data from an increasing number of IID spaces. 
Similar to Sec.~\ref{sec:role_of_mm}, we use ProcTHOR~\citep{procthor} to generate a large number of similar houses and pre-train models using an increasing number of them.
Fig.~\ref{fig:house-scaling} compares their performance on a held-out test space against a model pre-trained directly on that space. 
We find that \textit{even thousands of similar spaces are not enough to substitute pre-training on the test space}.

\begin{figure*}
    \centering
    \includegraphics[width=\textwidth]{figures/qual_fig_v3.pdf}
    \caption{\textbf{\tspmm predictions across different tasks}. We present qualitative results for \tspmm against various baselines, including scratch (no pre-training) and Internet-based pre-training on \textbf{real-world scenes} from Scannet++~\citep{yeshwanthliu2023scannetpp}. \tspmm predictions are notably more consistent across both tasks. Note how \tspmm predicts the same object (magnified in red boxes) more accurately and robustly across various viewpoints, as compared to generalist models like 4M-21~\citep{4m21} and DINOv2~\citep{oquab2023dinov2}. 
    }
    \vspace{-0.5em}
    \label{fig:maxmod-qual}
\end{figure*}

\noindent\textbf{Specialization-generalization trade-off.} 
We observed that the best performance on a given test space is achieved when pre-trained on data from the same space. 
Our cross-space analysis also shows that this specialization comes at the cost of performance on other spaces. 
Can we keep (or improve) this specialization performance while gaining generalization capabilities by adding more houses during pre-training in addition to the test house? We fix the model size and the total number of pre-training images, and progressively replace images from the test space with images drawn from an increasing number of similar houses. This keeps the overall training compute budget fixed, while varying the number of spaces represented in the pre-training data. 
Fig.~\ref{fig:spec-gen-trade-off} shows that as we add more houses during pre-training, the performance on the held-out new houses increases, as expected.
However, the performance on the original test space drops compared to the specialist test-space pre-training, demonstrating a \textit{specialization-generalization trade-off}.

\noindent\textbf{Additional results in \supmat.} 
Besides the analysis presented here, in the \supmat, we present more experiments on deploying \tsp in the wild (\appref\ref{appendix:deploy_wild}), \tsp using other self-supervised objectives (\appref\ref{appendix:other-ssl}), \tsp as label propagation (\appref\ref{appendix:label_prop}), the role of the transfer dataset mix-in during pre-training (\appref\ref{appendix:transfer-mix-in}), results for cross-modal retrieval (\appref\ref{appendix:cross-modal-retrieval}). We also present qualitative video evaluations on real-world spaces on our \href{https://tst-vision.epfl.ch/}{website}. 
Additionally, along with our code, we open-source a Swift-based iOS application (\appref\ref{appendix:iphone_app}) that leverages the ARKit API\footnote{https://developer.apple.com/documentation/arkit} to record the outputs of various sensors, including RGB, LiDAR, IMU, magnetometer, and ambient lighting.

\begin{figure*}[t!]
    \centering
    \includegraphics[width=0.8\textwidth]{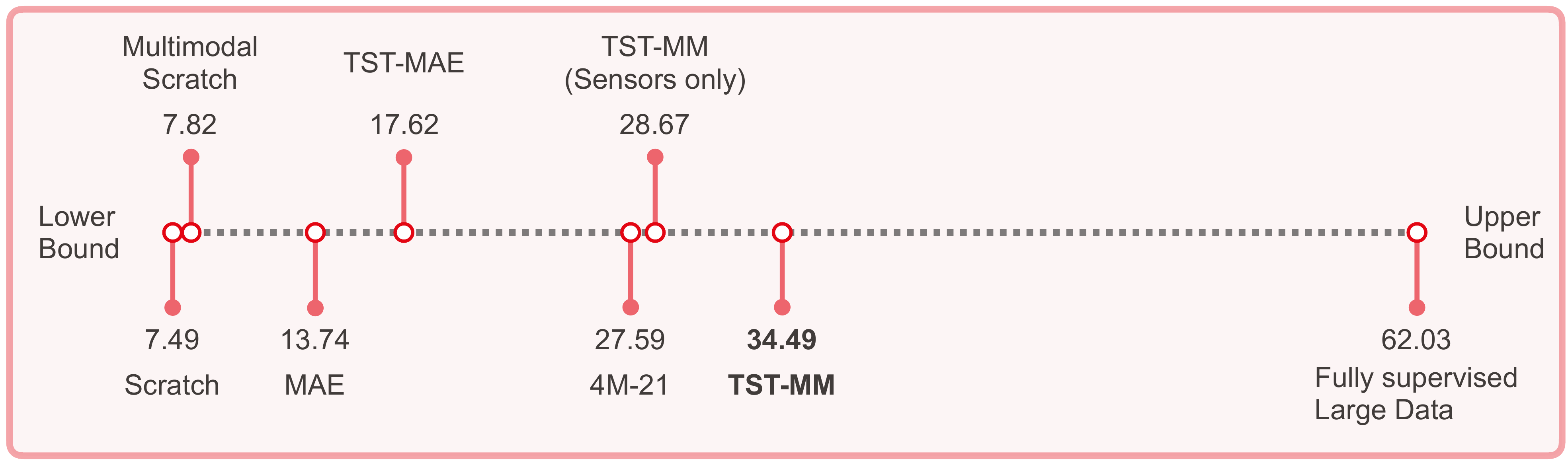}
    \vspace{-0.6em}
    \caption{\textbf{Quantitative summary of \tsp}.
    The lower bound is scratch (i.e., no pre-training and learning using the external transfer set only). The upper bound is approximated by a fully supervised model trained with a large number of annotated segmentation images. The gap between the lower and upper bounds is the playfield for the pre-training methods to fill. All methods share the same model architecture. The results here correspond to semantic segmentation on ScanNet++.
    }
    \vspace{-1em}
    \label{fig:tst_consolidated}
\end{figure*}

\subsection{Broader impact of \tsp}
\label{sec:obvious}
Our work sheds light on the critical role of data and data sources in building vision systems. The setting studied in \tsp
suggests that it is possible to achieve competitive results without relying on large, diverse internet-based datasets~\cite{Changpinyo2021CC12M,Schuhmann2022LAION5B} that essentially require the data of different users to be harvested and mixed. It
shows that training on only the deployment space data is an alternative worth considering and investigates
the requirements of making that viable (e.g., utilizing multimodality being critical for achieving good
results). Beyond performance, specialization can also improve deployment efficiency, since the one-time cost of test-space pre-training can be amortized when smaller specialist models replace larger generalists at inference (\appref~\ref{appendix:test_space_efficiency}).

This paradigm enables segregating the data of different users, which can enable a complete in-house, local training setup under full user control for privacy-critical scenarios, thereby avoiding any potential data contact with the external world. 
We provide an extended discussion on this in \appref~\ref{appendix:privacy_tst}. 

\section{Discussion and future work}
\label{sec:conclusion}

Our work presents a controlled study on leveraging multimodality as a source of self-supervision. 
We move away from pre-collected internet-based datasets to a setup where a multisensory agent moves around and collects multimodal data in a known deployment space. 
We find that, in this specialization setup, cross-modal learning results in effective representations from the test space and serves as a strong alternative to internet-based generalists~\citep{4m21,oquab2023dinov2}. 


We present a consolidated picture of our results in Fig.~\ref{fig:tst_consolidated} and highlight the following key takeaways:

\noindent$\bullet$ \textbf{\tspmm yields the most performant model in the test space.} It currently sits halfway between the bounds. It leverages cross-modal learning with sensory and off-the-shelf pseudolabelled modalities, with just test space data for pre-training. Note that this model does not qualify for claiming ``no external access''(Sec.~\ref{sec:tst_vs_internet}). 

\noindent$\bullet$ \textbf{Distilling pseudo-labeler outputs on test-space data is more effective than doing so on external data.} 
As discussed in Sec.~\ref{sec:tst_vs_internet}, using pseudo-label network outputs as modalities is equivalent to distilling their knowledge on test-space data. 
Comparing \tspmm with 4M-21~\citep{4m21} shows that doing this distillation on test-space data is more effective than doing so on external data (CC12M). Note that \tspmm and 4M-21 share the same architecture and pseudo-labelers. The same observation was consistently made for various settings in Fig.~\ref{fig:distillation} of the \appref~\ref{appendix:distillation}. This suggests that the benefit comes from distilling pseudo-label knowledge on test-space, rather than the pseudo-labelers alone. 


\noindent$\bullet$ \textbf{\tspmm (Sensors only) with no external access is competitive.} As discussed in Sec.~\ref{sec:howfar}, the \tspmm (Sensors only) uses no external data, directly or indirectly, as it only uses on-device sensors as modalities. It covers roughly 1/3 of the way, providing a nontrivial value, and is competitive with internet-based generalists such as 4M-21, trained on large-scale internet data (CC12M~\citep{Changpinyo2021CC12M}).

\noindent$\bullet$ \textbf{Multimodality as supervision is effective.} Across our results, multimodal models consistently outperform their single-modal counterparts (e.g., \tspmm vs \tsprgb or 4M-21~\citep{4m21} vs MAE~\citep{He2021MaskedAA}). Also, note that multimodality was necessary to make \tspmm a viable alternative to internet-based generalists~\citep{oquab2023dinov2,radford2021clip} in the test space, as \tsprgb performs suboptimally in comparison. 

Although cross-modal learning with multimodal masked modelling achieves state-of-the-art performance in the test space, there is still room for improvement (Fig.~\ref{fig:tst_consolidated}) to close the gap with the fully supervised upper bound. Future work can address this gap through stronger multimodal pre-training objectives and active data collection policies guided by the pre-training objective to select informative views of the test space. 
Other directions we are interested in include incorporating inductive biases such as multi-view consistency~\citep{luo2020consistent} and additional hardware sensory signals, e.g., IMU and magnetometer measurements.

\noindent\textbf{Acknowledgements.} This work was supported under project ID \textbf{43} as part of the Swiss AI Initiative, through a grant from the ETH Domain and computational resources provided by the Swiss National Supercomputing Centre (CSCS) under the Alps infrastructure. This material is based on work that is partially funded by an unrestricted gift from Google. This work has received funding from the Swiss State Secretariat for Education, Research and Innovation (SERI).
We also thank Daniel Filipe Jana and the EPFL \href{https://www.epfl.ch/research/facilities/scitas/}{SCITAS} team for their support. The authors also thank Chandan Yeshwanath for their help with the Scannet++~\citep{yeshwanthliu2023scannetpp} dataset, and Roman Bachmann for useful discussions.

{
    \small
    \bibliographystyle{iclr2026_conference}
    \bibliography{iclr2026_conference}
}

\input{tst_arxiv_suppmat}

\end{document}

%% file: math_commands.tex
\usepackage{amsmath,amsfonts,bm}

\def\eqref#1{equation~\ref{#1}}

\def\1{\bm{1}}

\DeclareMathAlphabet{\mathsfit}{\encodingdefault}{\sfdefault}{m}{sl}
\SetMathAlphabet{\mathsfit}{bold}{\encodingdefault}{\sfdefault}{bx}{n}



%% file: tst_arxiv_suppmat.tex
\clearpage
\appendix

\section*{\LARGE Appendix}

\startcontents[appendices]
\printcontents[appendices]{l}{1}{\setcounter{tocdepth}{2}}



\section{Additional related work.}
\label{appendix:additional_related_work}
\noindent\textbf{Domain Adaptation} in vision~\citep{Li2017DeeperBA,Zhou2021DomainGA} addresses the gap between a source domain, where abundant data is available, and the target domain, where limited~\citep{Shu2019TransferableCF,Liu2024UniversalAS} or no data~\citep{Dong2021WhereAH,Ganin2014UnsupervisedDA} are available. \tsp, when initialized from an Internet-based model, as presented in Tab.~\ref{tab:max-mod-quant}, can be seen as an instantiation of adapting a generalist model to the test space. However, \tsp differs by learning task-agnostic representations by self-supervised pre-training in the test space, as opposed to domain adaptation, which generally adapts a pre-trained task-specific network~\citep{Xu2021CDTransCT,Kang2019ContrastiveAN}.  

\noindent\textbf{Semi-Supervised Learning} refers to a line of work that attempts to learn a task from a limited labeled dataset and massive unlabeled data~\citep{vanEngelen2019ASO}. Clearly, it involves consistency regularization~\citep{Berthelot2019MixMatchAH,Sohn2020FixMatchSS,Xie2019UnsupervisedDA} and pseudo-labeling ~\citep{Guo2022RobustSL,Chen2021SemiSupervisedSS,Zhang2021FlexMatchBS,Liu2021UnbiasedTF} to generate supervision of unlabeled data, followed by joint training. Our framework, \tsp is closer in spirit to \textit{self-supervised learning}, as it tries to learn a task-agnostic representation for the test space, that we transfer for various downstream tasks like segmentation, detection and image captioning. Under semi-supervised learning, specialization with \tsp can be posed as using unlabelled data from the test space, as opposed to other sources like Internet or similar spaces. 

\noindent\textbf{Embodied Active learning.} In another line of work, SEAL~\citep{chaplot2021seal}, Interactron~\citep{kotar2022interactron}  learn a reinforcement learning-based policy to collect supervision in a house to finetune an off-the-shelf MaskRCNN~\citep{He2017MaskRCNN}, or observe additional frames for multi-frame inference for object detection. 
As opposed to focusing on adapting task-specific models, we focus on learning task-agnostic pre-trained representations over a test space. 

\section{Specialization Can Reduce Deployment Cost}
\label{appendix:test_space_efficiency}
\begin{wrapfigure}{r}{0.42\linewidth}
    \vspace{-1.0em}
    \centering
    \includegraphics[width=0.8\linewidth]{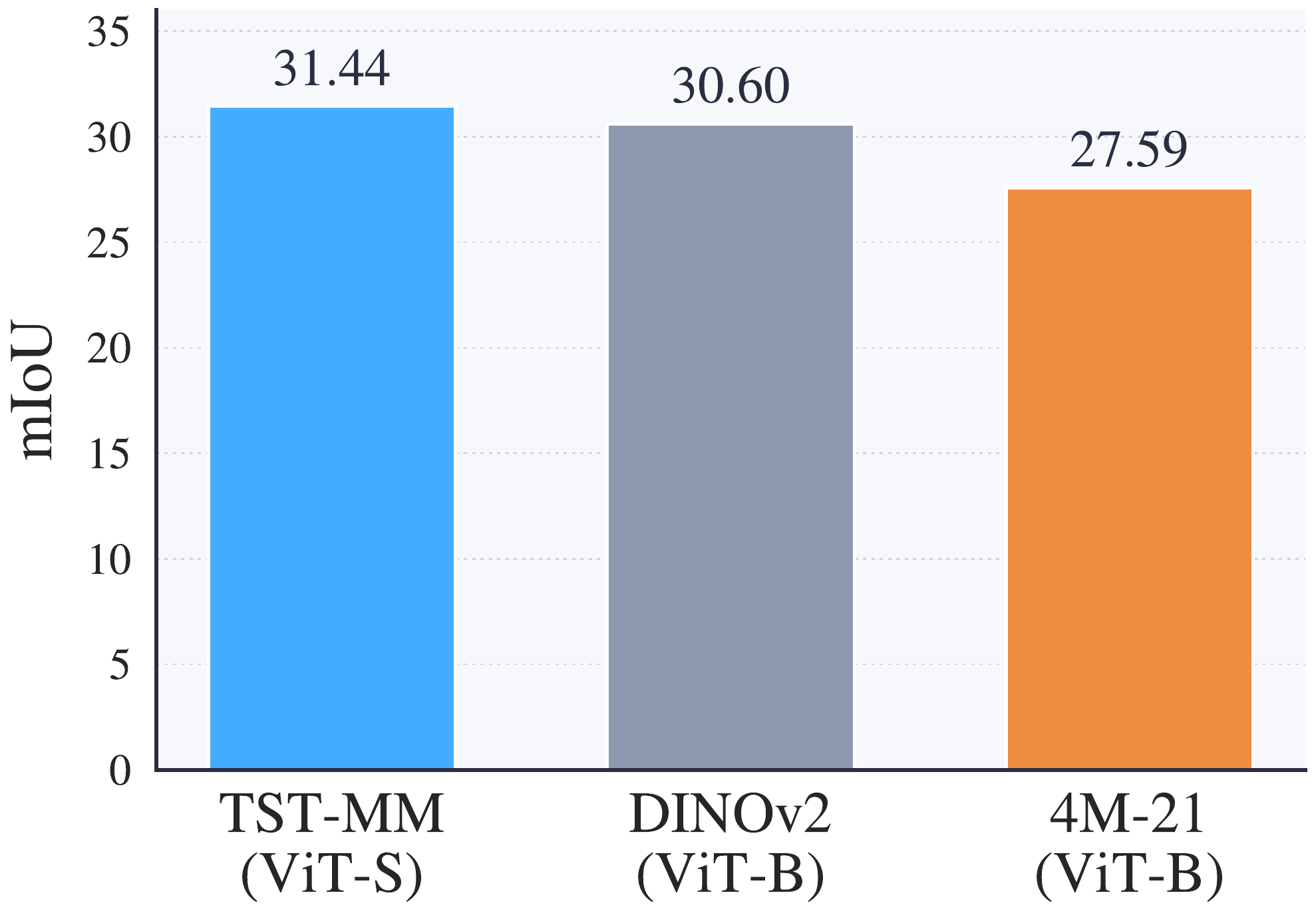}
    \caption{\textbf{\tsp can achieve similar performance with smaller model sizes.} We find that a specialized model with a ViT-S backbone can outperform internet-based generalist models such as DINOv2 and 4M-21 with a ViT-B backbone, which is 4x more computationally expensive to run inference on.}
    \label{fig:efficiency_plot}
    \vspace{-0.8em}
\end{wrapfigure}

We find that cross-modal learning on test-space data yields specialist models that can outperform internet-pretrained generalist models (Tab.~\ref{tab:max-mod-quant}). This specialization, however, introduces an additional cost: unlike generalist pre-training, which is performed once and reused across deployment spaces, \tsp requires pre-training a separate model for each target space, causing training compute to scale with the number of deployment spaces.

However, \tsp can also shift the efficiency tradeoff. As shown in Fig.~\ref{fig:efficiency_plot}, \tsp achieves strong performance with a smaller transformer backbone, partially offsetting the cost of deployment-specific pre-training through reduced inference compute. In particular, because ViT-S is $4\times$ faster at inference than ViT-B, the one-time cost of specialist pre-training can be amortized over repeated deployment: after a break-even point, serving a smaller specialist model becomes cheaper than serving a larger generalist model. This observation is consistent with recent work on specialist pre-training for language models~\citep{Baek2026TheFF}, which finds that smaller specialist models can be both cheaper and more performant to serve than large-scale generalist models.

\section{Label Propagation with \tsp}
\label{appendix:label_prop}
As mentioned in Sec.~\ref{sec:evaluation}, all results presented so far use external transfer images i.e., the annotated images used for training the transfers are from other spaces, rather than the test space. 
This choice matches a real-world setup, as it is infeasible for a user to spend time annotating images from their space. 
However, this introduces a minor suboptimality in quantifications, since the transfer set may exhibit distribution shifts from the test space (e.g., some test objects may not appear in the transfer set or have vastly different appearances).
This mismatch makes it difficult to attribute poor results primarily to ineffective pre-training. 

\begin{figure*}[t!]
    \centering
    \includegraphics[width=0.8\textwidth]{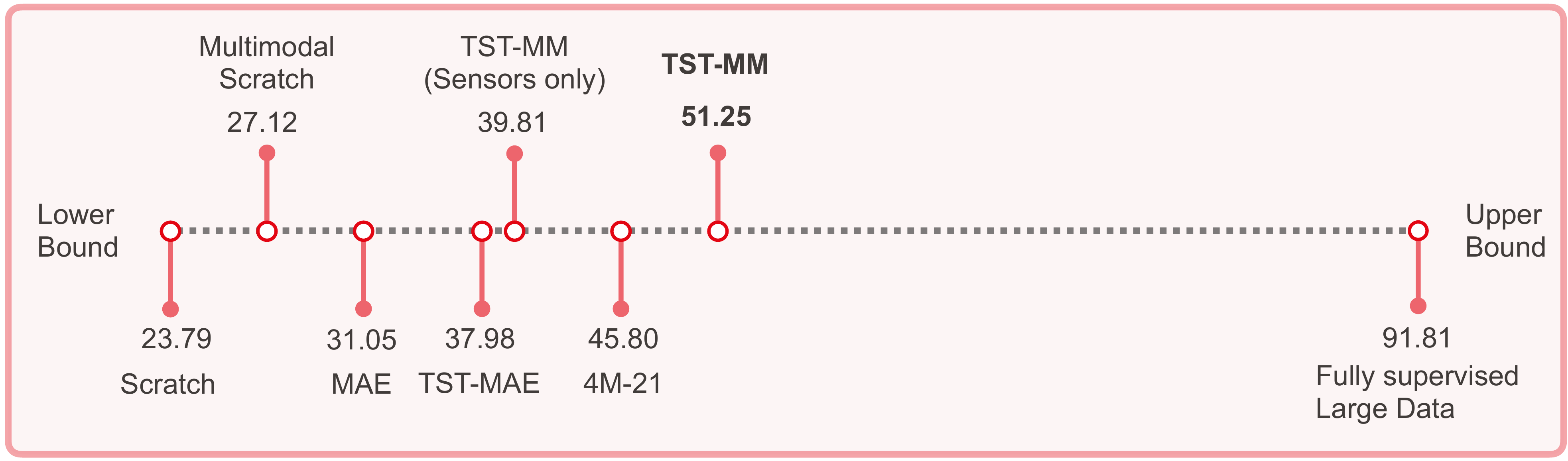}
    \caption{\textbf{Label Propagation with \tsp}.
    We find that the trends in performance under the Label Propagation, with no transfer set distribution shift, are similar to what we observe in Tab.~\ref{tab:max-mod-quant} and Fig.~\ref{fig:tst_consolidated}, thereby reassuring the efficacy of pre-training in the test space.
    The lower bound here is learning from scratch (i.e., no pre-training and learning using the small transfer set only). The upper bound is approximated by a fully supervised model trained with a large number of annotated segmentation images from the test space. 
    All methods share the same model architecture. 
    The results here correspond to semantic segmentation on Scannet++~\citep{yeshwanthliu2023scannetpp}. 
    }
    \label{fig:tst_consolidated_labelprop}
\end{figure*}

To alleviate this, we experiment with another setup where transfer images are a small set ($\approx$ 20) from the same test space. This is akin to having a few annotated images from the test space and expecting the model to propagate the labels to the rest of the images of the test space. Therefore, we name this setup as \textit{Label Propagation}.
This allows us to leverage this setup as a sandbox for studying various self-supervised pre-training methods in the test space. We present the results of various \tsp variants and relevant baselines under this setup in Fig.~\ref{fig:tst_consolidated_labelprop}.  

\section{Dataset Details.}
\label{appendix:dataset-details}
\noindent\textbf{1. Scannet++}~\citep{yeshwanthliu2023scannetpp} is a large dataset of real-world indoor spaces containing sub-millimeter resolution laser scans, paired with DSLR and iPhone RGB images.
\begin{itemize}
    \item \textbf{Pre-training dataset.} We use 8 Scannet++~\citep{yeshwanthliu2023scannetpp} scenes as our test space. We use a mix of iPhone and DSLR images for pre-training, with the iPhone containing $19165$ samples and the DSLR dataset containing $15000$ samples.
    \item \textbf{Transfer dataset.} We use non-test space buildings for creating a transfer set of $40000$ RGB, segmentation pairs. Note that Scannet++~\citep{yeshwanthliu2023scannetpp} only provides 3D instance annotations, which we project to 2D to create a semantic segmentation dataset. 
    \item \textbf{Evaluation.} We evaluate on semantic segmentation in the test space. The test dataset for evaluation contains $3000$ RGB image samples. Note that we collect a separate held out set from the test space for this stage. 
\end{itemize}

\noindent\textbf{2. Replica}~\citep{straub2019replica} provides high quality 3D reconstructions of real indoor spaces. 
\begin{itemize}
    \item \textbf{Pre-training dataset.} We use Omnidata~\citep{Eftekhar2021Omnidata}, to densely sample Replica meshes corresponding to the 5 scenes to build our pre-training dataset, $D_{PT}$, containing $84889$ samples. We defer the details of the sampling procedure to Omnidata~\citep{Eftekhar2021Omnidata}. 
    \item \textbf{Transfer dataset.} Similar to Scannet++~\citep{yeshwanthliu2023scannetpp}, we collect a transfer set from another set of Replica scenes that are different than the scenes used during pre-training. We collect $20000$ RGB images and semantic segmentation masks, and use it as our transfer dataset, $D_t$.
    \item \textbf{Evaluation.} We evaluate on semantic segmentation in the test space. We collect a test set of $5000$ images and semantic segmentation annotations from the same test space we pre-train on, and report performance on it. We leverage Omnidata annotation pipeline to extract the segmentation labels.    
\end{itemize}

\noindent\textbf{3. ProcTHOR}~\citep{procthor} It includes procedurally generated house-like environments. We use 5 procedurally generated houses as our test space. 
\begin{itemize}
    \item \textbf{Pre-training dataset.} We randomly sample various agent $x,y,z$ positions and orientations along its axis in the test space, and collect RGB-D images at these points. This sampling process yields a total of $163767$ samples. We collect data by sampling densely across the test space and use it as our pre-training dataset $D_{PT}$.
    \item \textbf{Transfer dataset.} For the transfer data $D_t$, we collect a small dataset of $20000$ RGB and task annotation pairs, from 800 houses generation using  a different asset and layout distribution than the pre-training test space, thereby making them out-of-distribution to it.  
    \item \textbf{Evaluation.} We evaluate \method and present results on three tasks, namely semantic segmentation, object detection and image captioning. We collect a test set with $5000$ samples from the same test space, where we performed pre-training, and report performance on it. We use the AI2-THOR~\citep{kolve2017ai2} metadata to extract semantic segmentation and object detection labels for evaluations. For captioning, we generate ground truth captions by prompting GPT-4o~\citep{openai2024hello} with privileged information, e.g. class names and bounding boxes.  Finally, we additionally evaluate our model on cross-modal retrieval~(in Sec.~\ref{appendix:cross-modal-retrieval}). 
\end{itemize}

\section{Additional downstream evaluations: Zero-shot cross-modal retrieval task}
\label{appendix:cross-modal-retrieval}

\begin{table}[h]
\centering
\begin{tabular}{@{}lcccccc@{}}
\toprule
\multicolumn{1}{c}{\multirow{2}{*}{Method}} & \multicolumn{3}{c}{Image to Depth}             & \multicolumn{3}{c}{Depth to Image} \\ \cmidrule(lr){2-4} \cmidrule(l){5-7} 
                                            & R@1 & R@5 & R@10 & R@1 & R@5 & R@10       \\ \midrule
4M-21~\citep{4m21}                           & 1.06                   & 2.18                   & 3.08                   & 1.0           & 2.76         & 3.66                       \\
\tspmm                                     & \textbf{25.48}           & \textbf{37.00}         & \textbf{41.58}    & \textbf{24.32}           & \textbf{36.46}         & \textbf{40.82}                          \\ \bottomrule
\end{tabular}%
\caption{\textbf{Zero-shot Cross-modal retrieval.} When performing the image-to-depth and depth-to-image cross-modal retrievals on the test space data using the predicted CLIP embeddings, we observe that the \tspmm method constantly outperforms the Internet-based 4M-21~\citep{4m21}.}
\label{tab:cross-modal-retreival}
\end{table}

\begin{figure*}
    \centering
    \includegraphics[width=0.9\textwidth] {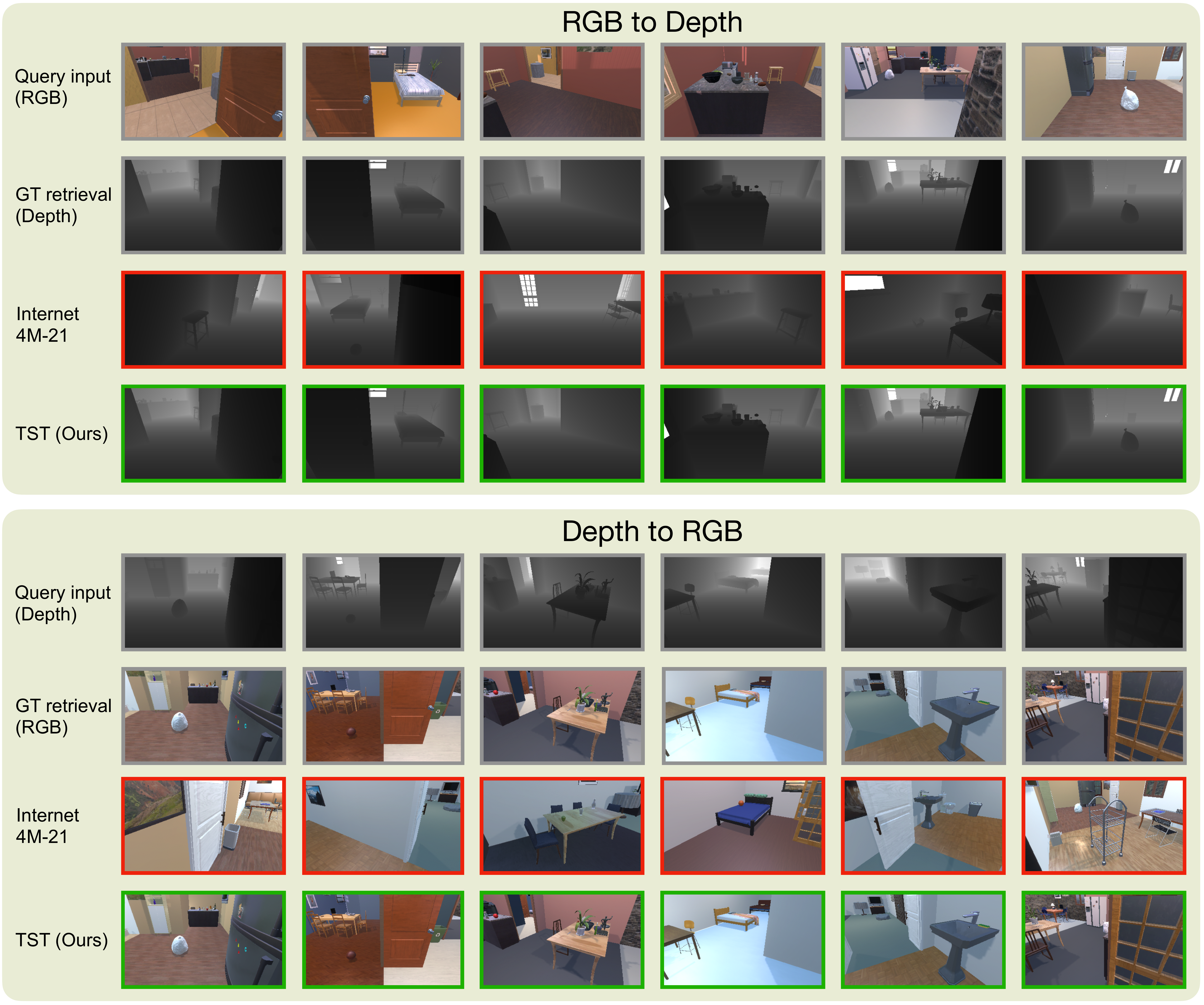}
    \caption{\textbf{\tspmm cross-modal retrieval predictions}. \tspmm retrieves corresponding RGB images from query Depth input and Depth images from RGB input more accurately than the Internet-based 4M-21\citep{4m21} model.}
    \label{fig:cross-modal}
    \vspace{-0.4em}
\end{figure*}

As mentioned in Section~\ref{sec:setup}, we present results on zero-shot cross-modal retrieval to further support our framework \tsp. Specifically, we evaluate the performance of models pre-trained with \tspmm on RGB-to-Depth and Depth-to-RGB retrieval. 
To perform retrieval using an Internet-based model, 4M-21~\citep{4m21} and \tspmm, we utilize their cross-modal generation capabilities by transforming depth and RGB images into CLIP embeddings, and then apply retrieval directly on the CLIP embeddings. 
Since 4M-21~\citep{4m21} and \tspmm generate feature maps for CLIP as the target modality from RGB and Depth images, we apply mean-pooling on the feature maps to obtain global CLIP embeddings. Cross-Modal retrieval evaluates \tspmm on two fronts: i) How well test-space paired modality inputs are aligned in the model representations internally, and ii) How effectively \tspmm can perform cross-modal generalization. For the evaluation, we report zero-shot recall at various thresholds on a test set of 5000 samples from ProcTHOR~\citep{procthor} test space.
The results are presented in Tab.~\ref{tab:cross-modal-retreival}. We also present qualitative examples in Fig.~\ref{fig:cross-modal}. Note that given our method \tspmm has access to the test space, it can retrieve RGB to Depth and Depth to RGB much more effectively than models based on external data like the Internet.


We find that \tspmm substantially outperforms 4M-21~\citep{4m21}. The recall performance of \tspmm further increases when evaluated on R@5 and R@10, whereas Internet-based 4M-21~\citep{4m21} shows diminishing returns. This underscores the effectiveness of test-space training, where specialization itself is crucial for learning test-space-aligned representations.

\section{\tspmm deployment in the wild.}
\label{appendix:deploy_wild}
\begin{wraptable}{r}{0.34\linewidth}
\vspace{-0.8em}
\centering
\begin{tabularx}{\linewidth}{l >{\raggedleft\arraybackslash}X}
\toprule
Method  & mIoU \\ \midrule
Scratch & 21.82 \\
4M-21   & 54.58 \\
\tspmm  & \textbf{59.11} \\ \bottomrule
\end{tabularx}
\caption{\textbf{TST deployment in the wild.} Semantic segmentation performance comparison across training methods.}
\label{tab:deploy_wild}
\vspace{-0.8em}
\end{wraptable}

\begin{figure*}
    \centering
    \includegraphics[width=0.70\textwidth] {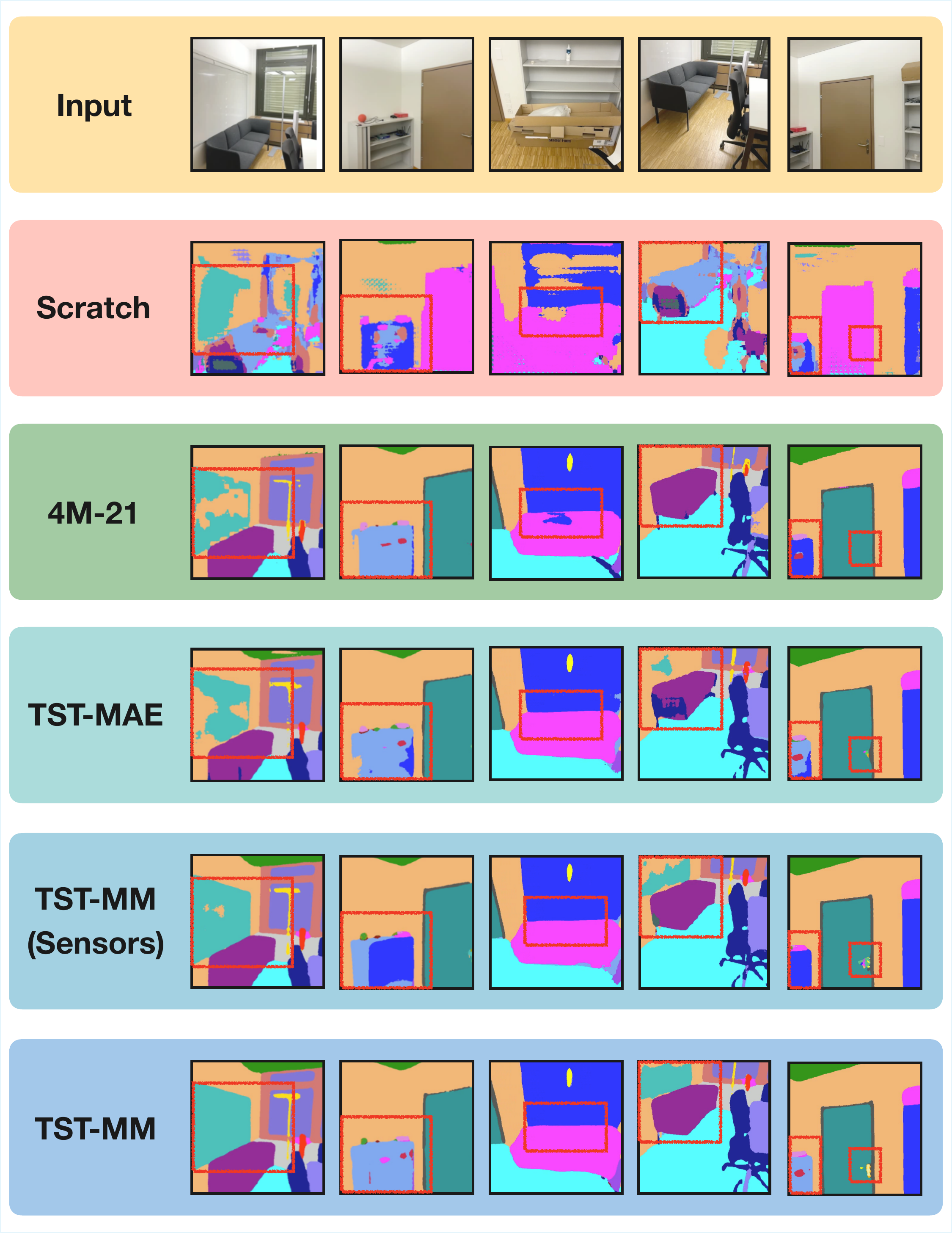}
    \caption{\textbf{\tspmm predictions on deployment in the wild}. We showcase the qualitative results for \tspmm on the semantic segmentation task against the Internet-based pre-trained model 4M-21\citep{4m21} and scratch (no-pretraining). \tspmm predictions are notably better across object categories, showing the value of access to test space and the deployment potential of \tspmm.}
    \label{fig:deploy_wild}
    \vspace{-0.4em}
\end{figure*}

In addition to real-world results on Scannet++~\ref{sec:tst_vs_internet}, we also experiment with the deployment of \tsp, in a custom space. We collect a 15-minute video of a meeting room and used the resulting frames for pre-training described in Sec.~\ref{sec:setup} followed by a transfer on the ScanNet++ \citep{yeshwanthliu2023scannetpp} transfer set (Sec. \ref{appendix:dataset-details}). We evaluated \tspmm and the baselines on the semantic segmentation task. We evaluate \tspmm and the baselines on the semantic segmentation task. Tab. \ref{tab:deploy_wild} shows that for this custom scene deployment, pre-training on the test-space through \tspmm outperforms the Internet-based baseline 4M-21 \citep{4m21}. The qualitative comparison in Fig. \ref{fig:deploy_wild} shows that \tspmm's predictions are notably better than those of the Internet-based 4M-21 \citep{4m21}. 



\section{\tsp with the DINOv2 objective.}
In this section, we explore how \tsp trained with DINOv2 objective, \tsp-\texttt{DINO} from \textit{scratch}, compares with its Internet counterpart trained on 142M images from the Internet~\citep{oquab2023dinov2}.
Fig.~\ref{fig:tst-dino} shows that pre-training on only data from the test space can substitute large-scale Internet pre-training.
This further underscores that \tsp framework extends to other self-supervised objectives~\citep{oquab2023dinov2} beyond masked modeling for specialization.
However, we find that \tspmm, which uses multimodal masked modeling outperforms other unimodal self-supervised objectives like DINOv2~\citep{oquab2023dinov2} and MAE~\citep{He2021MaskedAA}.

\section{Benchmarking different self-supervised objectives under \tsp.}
\label{appendix:benchmark_ssl_obj}
\begin{figure*}[t]
    \centering
    \begin{minipage}[t]{0.48\textwidth}
        \centering
        \includegraphics[width=\linewidth]{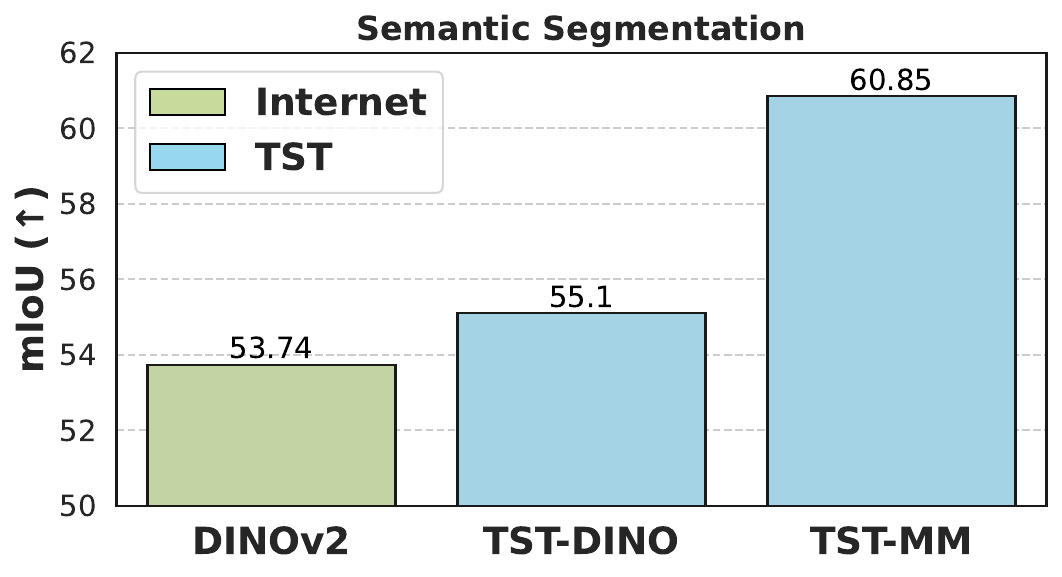}
        \captionsetup{type=figure}
        \caption{
        \footnotesize
        \textbf{\tsp with DINOv2 objective outperforms its Internet counterpart.}
        We compare the performance of DINOv2 pre-training in the test space, \tsp-\texttt{DINO}, with DINOv2 pre-trained on the large-scale Internet dataset of 142M images~\citep{oquab2023dinov2}.
        \tsp-\texttt{DINO} outperforms its Internet counterpart, showing the value of specialization.
        Yet, \tspmm with multimodal masked modeling achieves the best performance.
        }
        \label{fig:tst-dino}
    \end{minipage}\hfill
    \begin{minipage}[t]{0.48\textwidth}
        \centering
        \includegraphics[width=\linewidth]{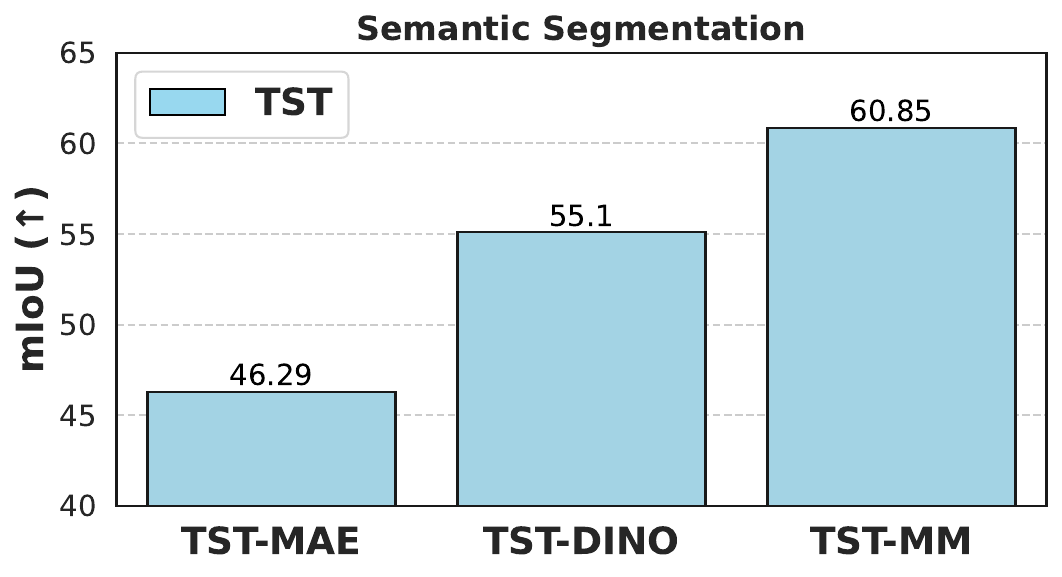}
        \captionsetup{type=figure}
        \caption{
        \footnotesize
        \textbf{Comparison between different pre-training objectives under the \tsp framework.}
        We compare the performance of different pre-training objectives using \tsp on the semantic segmentation task. We find that multimodal masked modeling (\tspmm) achieves the best performance followed by \tsp-\texttt{DINO}. All the three objectives were trained using the ViT-B model size on the ProcTHOR~\citep{procthor} dataset.
        }
        \label{fig:tst-obj-comparison}
    \end{minipage}
\vspace{-0.5em}
\end{figure*}

We compare the performance of \tsp with different pre-training objectives such as multimodal masked modeling~\citep{4m21}, DINOv2~\citep{oquab2023dinov2} and MAE~\citep{He2021MaskedAA}. As shown in Fig.~\ref{fig:tst-obj-comparison}, we find that multimodal masked modeling (\tspmm) to be the most performant among the self-supervised objectives we explored. However, note that all 3 objectives show specialization trends as presented in Fig.~\ref{fig:1h-confusion} and Fig.~\ref{fig:1h-confusion-rhs}.

\begin{figure}
    \centering
    \includegraphics[width=\columnwidth]{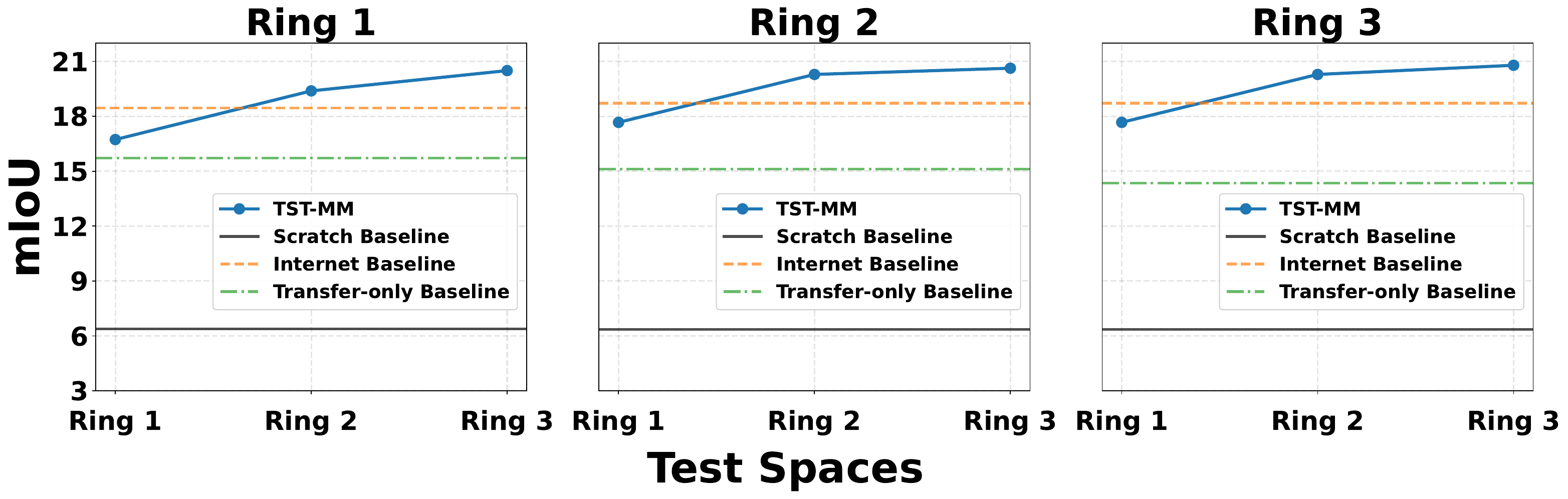}
    \caption{\textbf{Smallest unit of space to specialize on.} We reduce the test space size, that we can specialize and pre-train models with \tspmm. We compare it with an Internet pre-trained model~\citep{4m21}, and a baseline that pre-trains only on the transfer set. We also find that training on a ring smaller than the test ring, leads to diminished performance.}
    \label{fig:rings}
\end{figure}

\section{What is the smallest unit of space we can specialize on?}
In the results presented so far, we have shown that \tsp can specialize on test spaces at the size from 1-8 houses. However, this raises a question, what is the smallest unit of space we can specialize on? To probe this, we reduce the size of the test space and evaluate if \tsp can specialize to it. We consider a model trained via \tsp specialized, if it can outperform an off-the-shelf Internet-based generalist, when evaluated on that test space. We reduce the test space, in the form of concentric rectangles, starting with a room, and then reducing the size of the rectangle. For each rectangle, we pre-train a specialist model via \tsp. We compare this against 4M-21~\citep{4m21}, on the task of semantic segmentation. As shown in Fig.~\ref{fig:rings}, we find that we can specialize on a single room (ring 3) that has an area of 20 square metres, and this trend continues as we reduce it down to ring 2, which is 11 square metres and ring 1 which is 5 square metres. Reducing the test space, below 5 square metres results in failed specialization, where the pre-training on just the transfer pre-training performs the best.

\begin{figure*}[t]
    \centering
    \begin{minipage}[t]{0.48\textwidth}
        \centering
        \includegraphics[width=\linewidth]{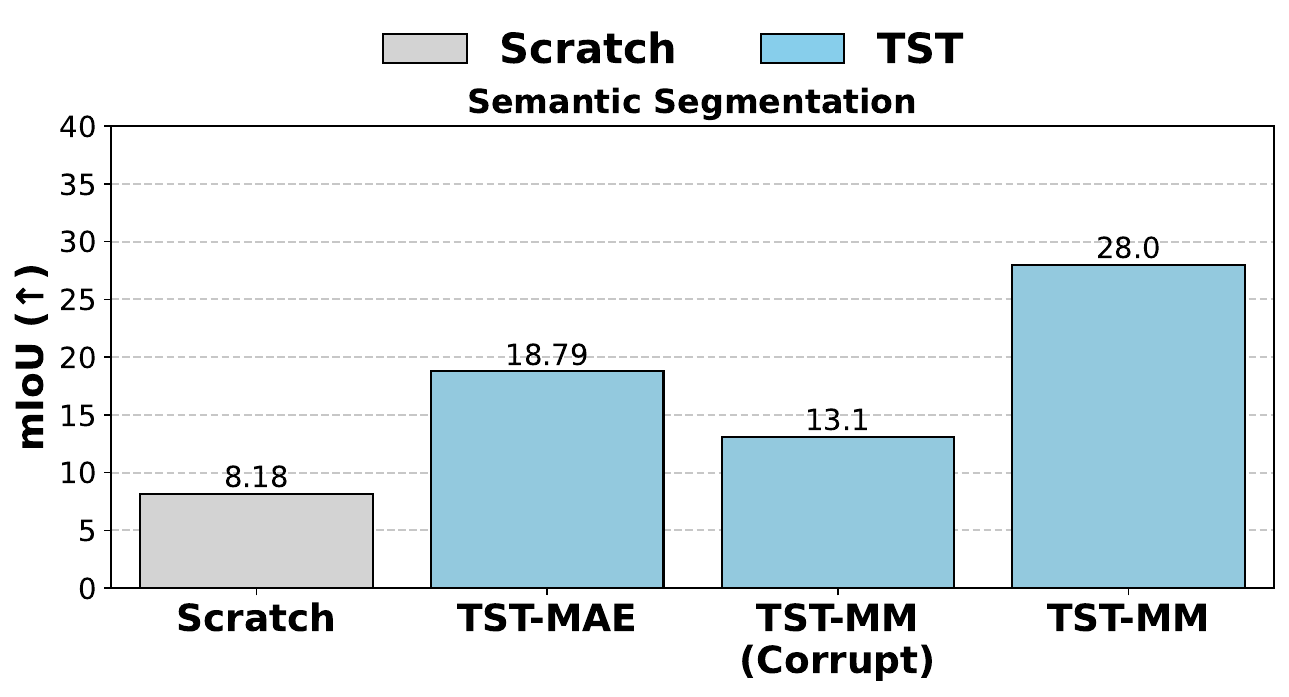}
        \captionsetup{type=figure}
        \caption{\textbf{\tsp with \textit{untrained} pseudolabel modalities.} We train variants of \tsp, with RGB-only (\tsprgb), with RGB and 4 pseudolabel modalities (Depth, Surface Normal, CLIP and Imagebind), with trained networks (\tspmm), and with untrained or \textit{corrupt} networks (\tspmm (Corrupt)). All results use a ViT-S backbone model on Scannet++.}
        \label{fig:untrained_modality}
    \end{minipage}\hfill
    \begin{minipage}[t]{0.48\textwidth}
        \centering
        \includegraphics[width=\linewidth]{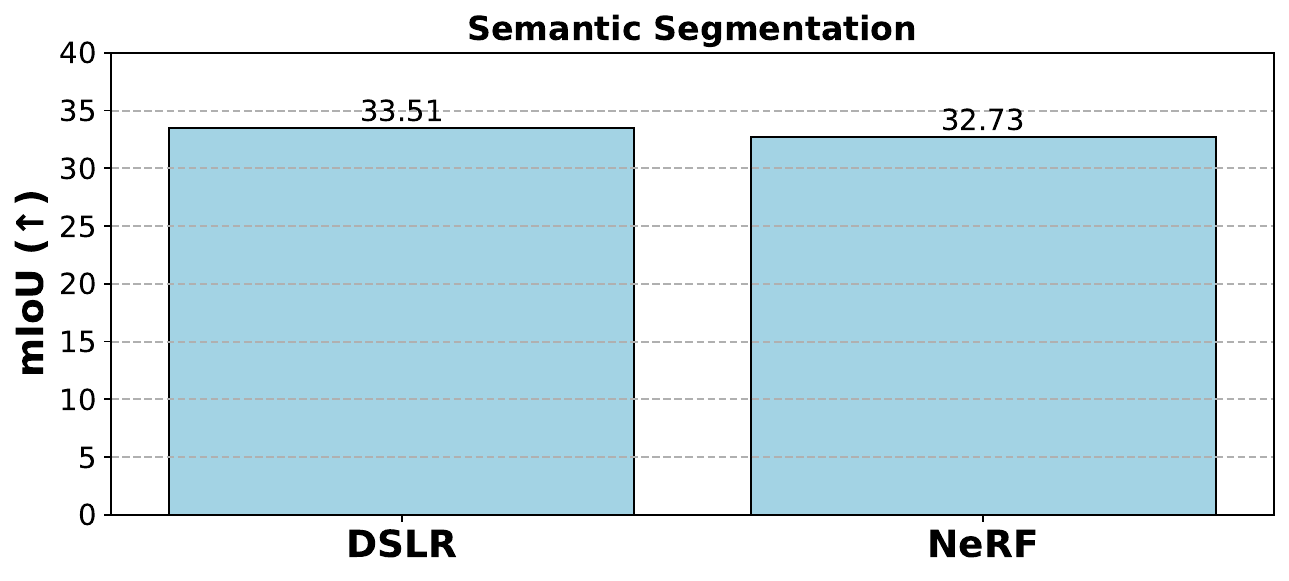}
        \captionsetup{type=figure}
        \caption{\textbf{\tsp with synthetic data.} We replace real DSLR images in ScanNet++~\citep{yeshwanthliu2023scannetpp} with NeRF~\citep{Mildenhall2020NeRF}-rendered images from the same training viewpoints. We find only negligible performance loss, demonstrating that NeRF output quality at known poses is sufficient to substitute high-quality DSLR images.}
        \label{fig:nerf_dslr}
    \end{minipage}
\end{figure*}

\section{Do we need \textit{trained} networks outputs for modalities?}
\label{appendix:untrained_pseudolabel}
As described in Sec.~\ref{sec:tst_vs_internet}, \tspmm leverages off-the-shelf pseudolabel networks to create additional modalities. This is akin to distilling their feature space on test space data, without directly accessing the external data that these networks were trained on. 
This raises an intriguing question, is it the trained feature space of a network~\cite{radford2021clip,Girdhar2023ImageBindOE}, that makes it a good modality, or simply the architecture inductive bias that creates a different transformation of the input image is sufficient to drive performance. To study this we present an analysis in Fig.~\ref{fig:untrained_modality}, which shows that using untrained pseudolabels as modalities drops performance below the RGB-only baseline, suggesting that model capacity spent on learning those modalities does not lead to a useful representation.Thi suggests that \textit{the learned feature space of pseudolabel network is crucial in making it as useful modality}. 

\section{\tsp with synthetic data.}
Recent advances in novel view synthesis~\citep{Mildenhall2020NeRF,Barron_2021_ICCV,Kerbl20233DGS} have enabled realistic renderings of indoor spaces, opening up the potential for generating synthetic training data. In \tsp, we leverage existing indoor scene datasets~\citep{yeshwanthliu2023scannetpp,straub2019replica}, which include real RGB images captured with DSLR/iPhone cameras or rendered from 3D meshes, to develop specialized models for specific test spaces.
This leads to a key question: if a novel view synthesis model can generate images from arbitrary viewpoints in a test space, can it serve as a controllable data generator—and can its outputs match real images in utility?

To explore this, we train a NeRF model~\citep{nerfstudio} using DSLR images from ScanNet++\citep{yeshwanthliu2023scannetpp}, and render images from the same camera poses. We then pre-train two models—one using real DSLR images and the other using NeRF-rendered views—to assess the performance gap. As shown in Fig.\ref{fig:nerf_dslr}, NeRF-generated images result in negligible performance loss compared to real images. This suggests an interesting future direction: if high-fidelity NeRF models can be trained with fewer input images, they could act as steerable data generators, reducing the need for extensive real-world data collection in test environments.

\begin{figure*}[t]
    \centering
    \begin{minipage}[t]{0.48\textwidth}
        \centering
        \vspace{-6.0em}
        \resizebox{\linewidth}{!}{
            \begin{tabular}{@{}lccc@{}}
                \toprule
                                         & Test Space & Transfer & Segmentation (mIoU~$\uparrow$) \\ \midrule
                                         & \xmark     & \cmark   & 42.01                         \\ \midrule
                \multirow{2}{*}{\tsp}  & \cmark     & \xmark   & \underline{50.21}             \\
                                         & \cmark     & \cmark   & \textbf{56.96}                \\ \midrule
                4M-21                    &            &          & 46.12                         \\ \bottomrule
            \end{tabular}
        }
        \captionsetup{type=table}
        \caption{\textbf{Ablating the use of transfer RGB frames during pre-training}. As noted in Sec.~\ref{sec:setup}, we additionally use RGB images from the transfer set during pre-training. We ablate this choice by comparing all three dataset configurations. We use the ViT-S backbone for all models.}
        \label{tab:test-transfer-ablation}
    \end{minipage}\hfill
    \begin{minipage}[t]{0.48\textwidth}
        \centering
        \includegraphics[width=\linewidth]{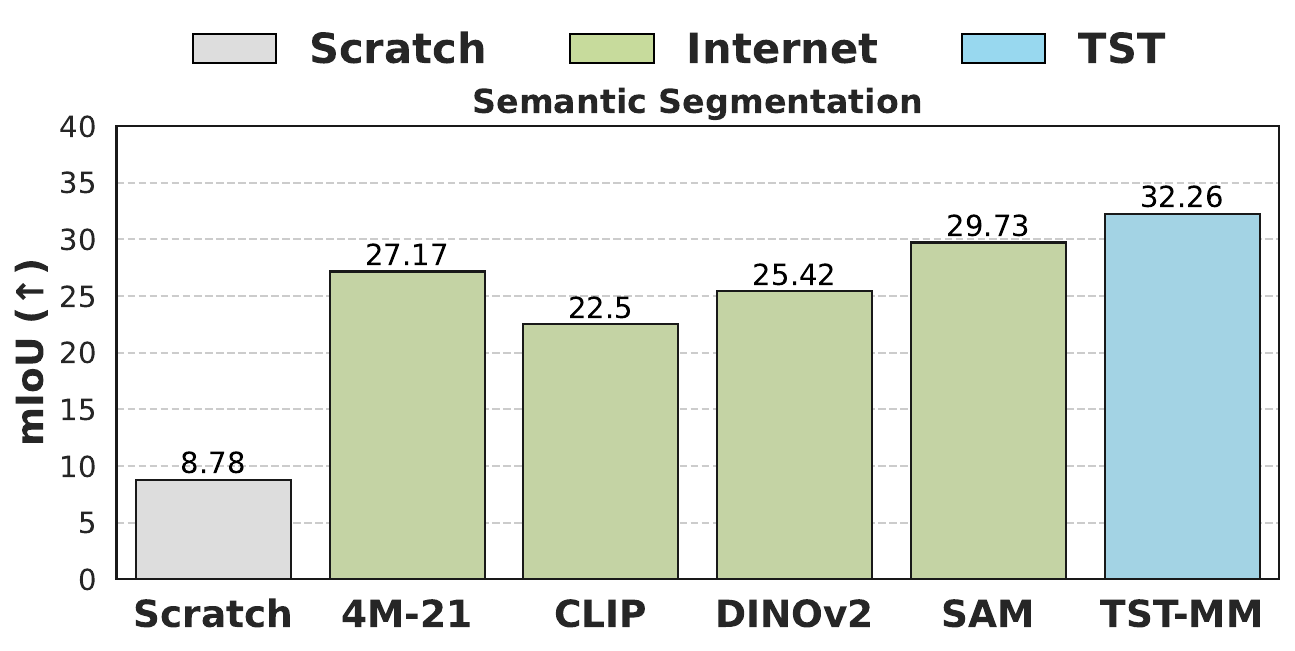}
        \captionsetup{type=figure}
        \caption{\textbf{\tsp works with off-the-shelf transfer set.} For Replica~\citep{straub2019replica}, we find that even when we use ADE20k~\citep{Zhou2017ADE20K} as a transfer set, \tspmm outperforms Internet-based generalist models, showcasing the importance of having access to the test space, agnostic to the transfer set.}
        \label{fig:cross_domain}
    \end{minipage}
\end{figure*}

\section{The role of the transfer dataset mix-in during pre-training.}
\label{appendix:transfer-mix-in}
We study the role of mixing images from the test space and transfer datasets during pre-training, as mentioned in Sec.~\ref{sec:setup}. 
Tab.~\ref{tab:test-transfer-ablation} shows that using only test-space data outperforms both pre-training on large-scale Internet data and using only transfer images, but mixing test space and transfer data achieves the best performance.
We hypothesize that seeing transfer images during pre-training helps the model to better align with the fine-tuning stage on the transfer dataset.
Note that it cannot be explained by more diverse data in the transfer set, as adding non-test spaces decreases the specialization performance, as observed in Fig.~\ref{fig:spec-gen-trade-off}.

\section{\tsp with off-the-shelf transfer set.}
As noted in Sec.~\ref{sec:setup}, for each dataset~\citep{procthor,straub2019replica,yeshwanthliu2023scannetpp} we explore, the transfer set comes from a similar domain, as the pre-training dataset, albeit from non-test spaces. It naturally raises the question, what if we use an existing off-the-shelf semantic segmentation dataset like ADE20k~\citep{Zhou2017ADE20K} as a transfer set. Does \tsp generalize and result in performant specialist models, or is an in-domain transfer set necessary? To probe this, for the Replica~\citep{straub2019replica} dataset, we pre-train \tspmm, but instead of using non-test spaces from Replica as the transfer set, we use ADE20k~\citep{Zhou2017ADE20K}. Fig.~\ref{fig:cross_domain} shows \tspmm outperforms various generalist models~\citep{4m21,oquab2023dinov2,radford2021clip}, even when using ADE20k~\citep{Zhou2017ADE20K} as the transfer set. All models are evaluated in the test space from Replica~\citep{straub2019replica}, on semantic segmentation, with a ViT-B backbone.

\section{Pseudo-labeler baselines}
\label{sec:pseudolabel_baseline}
As mentioned in Sec.~\ref{sec:tst_vs_internet}, we use various off-the-shelf networks to pseudolabel RGB data, and create additional (optional) modalities for \tspmm. We present a comparison for \tspmm against these pseudolabel baselines in Tab.~\ref{tab:ps-quant}. \tspmm and \tspmm (adapted) outperform all pseudolabel baselines, suggesting the benefit of pre-training in the test space with them, via multimodal masked modeling.

\begin{table*}[t!]
\centering
\resizebox{\linewidth}{!}{%
\begin{tabular}{llccccccc}
\toprule
 & \multirow{3}{*}{Method} & \multicolumn{3}{c}{Semantic Segmentation} & \multicolumn{2}{c}{Object Detection} & \multicolumn{2}{c}{Captioning} \\ 
\cmidrule(lr){3-5} \cmidrule(lr){6-7} \cmidrule(lr){8-9} 
 & & Scannet++ & ProcTHOR & Replica & Scannet++ & ProcTHOR & \multicolumn{2}{c}{ProcTHOR} \\ 
 & & mIoU $\uparrow$ & mIoU $\uparrow$ & mIoU $\uparrow$ & mAP $\uparrow$ & mAP $\uparrow$ & CIDEr $\uparrow$ & SPICE $\uparrow$ \\ 
\midrule
\multirow{5}{*}{\shortstack[l]{Pseudo-labelers}} 
 & ImageBind~\citep{Girdhar2023ImageBindOE} & 25.40 & 44.54 & 12.78 & 6.78 & 32.54 & - & - \\ 
 & CLIP~\citep{radford2021clip} & 23.02 & 48.66 & 20.92 & 19.75 & 38.47 & 18.4 & 16.2 \\ 
 & Mask2Former~\citep{cheng2021mask2former} & 29.42 & 50.28 & 22.68 & - & - & - & - \\ 
 & ViTDet~\citep{li2022vitdet} & - & - & - & 23.49 & 44.10 & - & - \\ 
 & SAM~\citep{Kirillov2023SAM} & \underline{34.75} & 56.72 & 28.51 & - & - & - & - \\ 
\midrule
\multirow{2}{*}{\shortstack[l]{Specialist \\ Pre-training}}                
& \tspmm                                      & 34.49     & \textbf{60.85} & \underline{32.87} & \underline{31.54}    & \underline{49.38}   & \underline{34.3}           & \underline{20.4}          \\
& \tspmm (adapted)                            & \textbf{36.44}         & \underline{60.59}              & \textbf{34.53}              & \textbf{35.83}                 & \textbf{51.25}                & \textbf{39.9}              & \textbf{20.5}             \\ 
\bottomrule
\end{tabular}
}
\caption{\textbf{Comparing \tspmm against pseudolabels.} We find that \tspmm outperforms all pseudolabels, underscoring the value of pre-training on them via multimodal masked modelling in the test space.}
\label{tab:ps-quant}
\end{table*}

\section{\tsp with no semantic modalities}
\label{appendix:baselines-additional}
\tspmm includes modalities obtained as outputs from different off-the-shelf models.
Tab.~\ref{tab:max-mod-quant} shows that \tspmm outperforms each individual model used as a modality.
Since our transfer tasks are semantic segmentation and object detection, we further study if having off-the-shelf models trained on related tasks as modalities is crucial for our final performance.

We present three experiments using the ViT-B backbone on ProcTHOR~\citep{procthor}.
For each experiment, we drop one of the following modalities: i) Semantic segmentation, ii) Object detection, iii) Semantic segmentation, Object detection, and SAM edges. Tab.~\ref{tab:semantic-drop-experiments} shows the results for each model when transferred to semantic segmentation and object detection. We find that even though the performance drops if we remove all three modalities, \tspmm still outperforms the Internet-based 4M-21~\citep{4m21} model.

\begin{table}[ht]
\centering
\resizebox{1\columnwidth}{!}{
\begin{tabular}{@{}lccccccc@{}}
\toprule
\multirow{2}{*}{Method} & \multicolumn{4}{c}{Modalities}                                                                                                                                                                             &  & \multicolumn{2}{c}{Task}                    \\ \cmidrule(lr){2-5} \cmidrule(l){7-8} 
                        & \begin{tabular}[c]{@{}c@{}}Semantic \\ segmentation\end{tabular} & \begin{tabular}[c]{@{}c@{}}Object \\ detection\end{tabular} & \begin{tabular}[c]{@{}c@{}}SAM\\ edge\end{tabular} & Others               &  & \begin{tabular}[c]{@{}c@{}}Segmentation \\ (mIoU$\uparrow$)\end{tabular}         & \begin{tabular}[c]{@{}c@{}}Detection \\ (mAP$\uparrow$)\end{tabular}             \\ \midrule
\multirow{4}{*}{\tspmm}&                                                                  \cmark&                                                             \cmark&                                                    \cmark&                      \cmark&  &                      \textbf{60.85}&                      49.38\\
                        &                                                                  \xmark&                                                             \cmark&                                                    \cmark&                      \cmark&  &                      59.43&                      \textbf{49.58}\\
                        &                                                                  \cmark&                                                             \xmark&                                                    \cmark&                      \cmark&  &                      59.38&                      49.34\\
                        & \multicolumn{1}{c}{\xmark}                                             & \multicolumn{1}{c}{\xmark}                                        & \multicolumn{1}{c}{\xmark}                               & \multicolumn{1}{c}{\cmark} &  & \multicolumn{1}{c}{55.39} & 45.97 \\ \midrule
 4M-21~\citep{4m21} & \cmark& \cmark& \cmark& \cmark& & 53.24 & 41.43\\ \bottomrule
\end{tabular}
}
\vspace{0.5em}
\caption{\textbf{The effect of semantic modalities in \tspmm.} As the results demonstrate, removing the semantic segmentation and object detection modalities obtained from off-the-shelf networks does not significantly hurt the \tspmm's performance on the downstream semantic segmentation and object detection tasks. When all three semantic modalities are removed, we observe a drop in performance, but \tspmm still outperforms the Internet-based 4M-21~\citep{4m21} model, demonstrating the value of specialization.}
\label{tab:semantic-drop-experiments}
\end{table}




\section{Do other self-supervised objectives benefit from specialized pre-training?}
\label{appendix:other-ssl}
\begin{wrapfigure}{r}{0.48\linewidth}
    \vspace{-3.0em}
    \centering
    \includegraphics[width=0.9\linewidth]{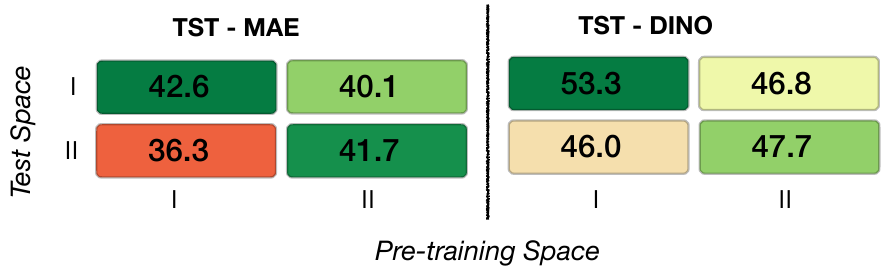}
    \vspace{-0.5em}
    \caption{\textbf{Specialization using other objectives.} We demonstrate specialization using other pre-training objectives, including MAE and DINOv2.}
    \label{fig:1h-confusion-rhs}
    \vspace{-0.8em}
\end{wrapfigure}
In Sec.~\ref{sec:tst_vs_internet}, we present results with \tspmm, which employs multimodal masked modeling. However, as mentioned in Sec.~\ref{sec:crossmodal}, \tsp also supports other self-supervised objectives. Fig.~\ref{fig:1h-confusion-rhs} shows that pre-training objectives, DINOv2~\citep{oquab2023dinov2}, and RGB-only MAE~\citep{He2021MaskedAA} exhibit similar specialization trends.

\section{Is distilling in the test space beneficial?}
\label{appendix:distillation}
As discussed in Sec.~\ref{sec:tst_vs_internet}, we scale modalities by pseudo-labelling RGB data with various Internet-based models~\citep{oquab2023dinov2,radford2021clip,cheng2021mask2former}. This process of creating additional modalities, and pre-training on them has enabled powerful multimodal foundation models~\citep{4m,4m21,bachmann2022multimae}. This form of pre-training can also be seen as distilling the knowledge from these powerful off-the-shelf networks into a single unified model. 

\begin{wrapfigure}{r}{0.46\linewidth}
\vspace{-1.0em}
    \centering
    \includegraphics[width=\linewidth]{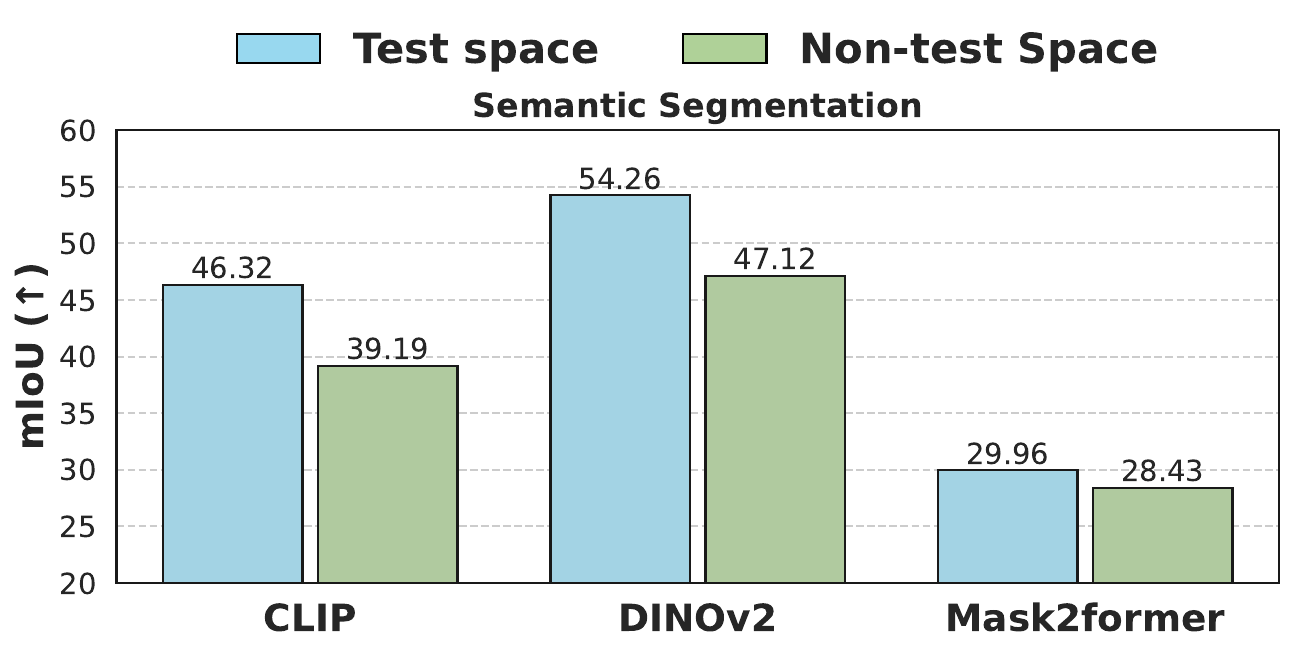}
    \caption{\textbf{Distillation in test space.} We find distilling over data from the test space, from various off-the-shelf models, results in more performant models in the test space. All results here are with the ViT-B backbone, on ProcTHOR~\citep{procthor}.}
    \label{fig:distillation}
\vspace{-0.8em}
\end{wrapfigure}
With \tspmm, we also distill from various off-the-shelf networks like CLIP~\citep{radford2021clip},  DINOv2~\citep{oquab2023dinov2} with masked modelling~\citep{He2021MaskedAA,4m}. Results in Tab.~\ref{tab:max-mod-quant}, suggest that distilling with multiple modalities on the test space, results in performant specialist models. However, to disentangle the effect of multimodality and distillation, we take it one step further to probe whether just distilling in the test space, provides some additional benefit, over non-test spaces? Therefore, we distill, CLIP~\citep{radford2021clip}, DINOv2~\citep{oquab2023dinov2} and Mask2former~\citep{cheng2021mask2former} in the test space, and compare it with distilling in an IID, but non test space, and report the results in Fig.~\ref{fig:distillation} on semantic segmentation in ProcTHOR~\citep{procthor}. We find that distilling over data from the test space is more performant than the data from non-test spaces, underscoring the importance of access to the test space for specialization.   


\begin{figure}
    \centering
    \includegraphics[width=0.8\columnwidth]{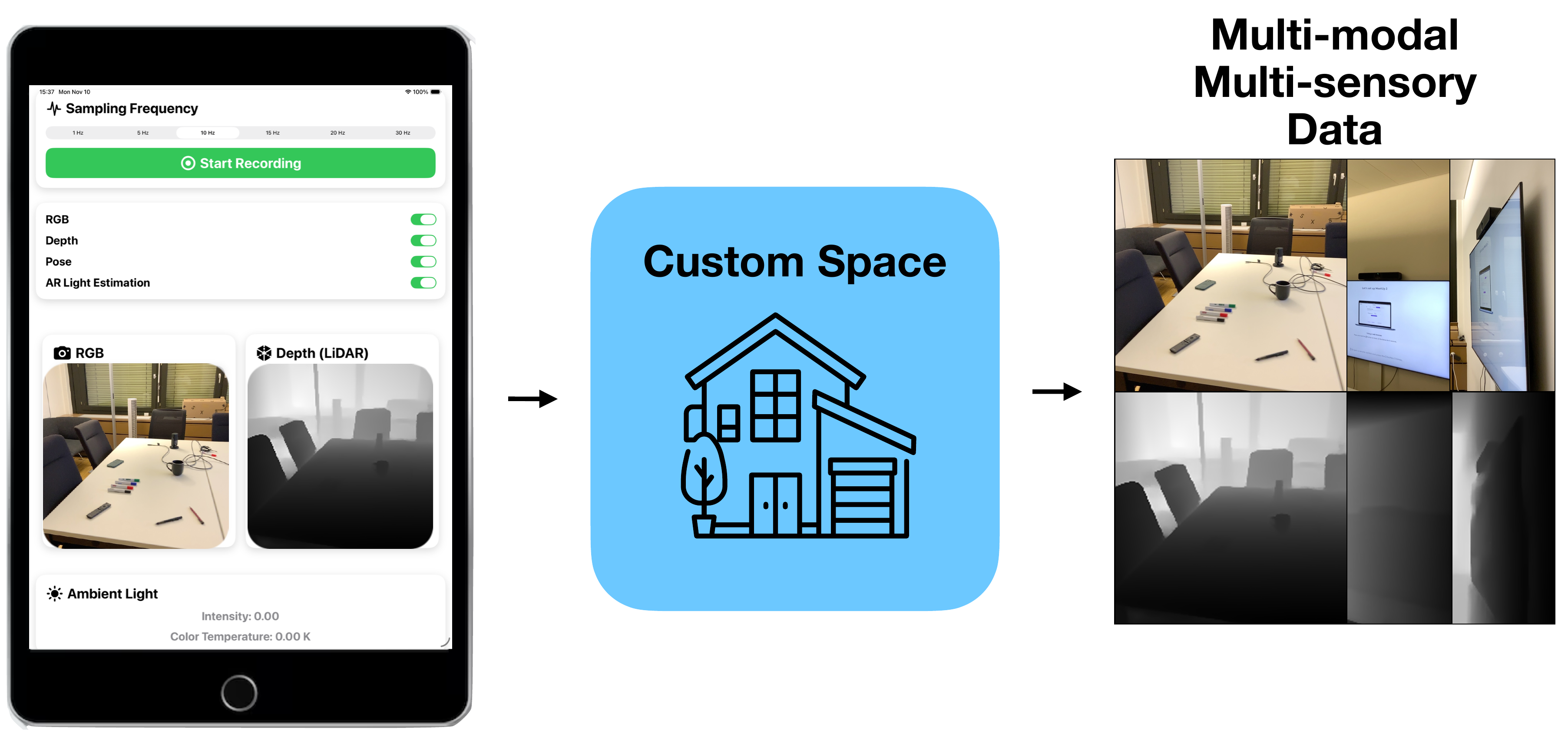}
    \caption{\textbf{iOS application for custom data collection.} It leverages the open source ARKit API to stream outputs of various sensors such as RGB, LiDAR, IMU, magnetometer, and
ambient lighting, and supports paired data collection.}
    \label{fig:arkit_app}
\end{figure}

\section{Application for hardware data collection}
\label{appendix:iphone_app}
As discussed in Sec.~\ref{sec:data}, \tsp can be extended to leverage more hardware-based modalities, such as IMU, GPS, Audio, which can be found on most common user devices, such as iPhone. To facilitate future research in this area, we release an iOS application that enables anyone to collect aligned multimodal data from RGB-D and additional hardware sensors present on an iPhone. Fig.~\ref{fig:arkit_app}, shows an overview of our application.

\begin{figure*}[t!]
\centering
\includegraphics[width=0.7\linewidth]{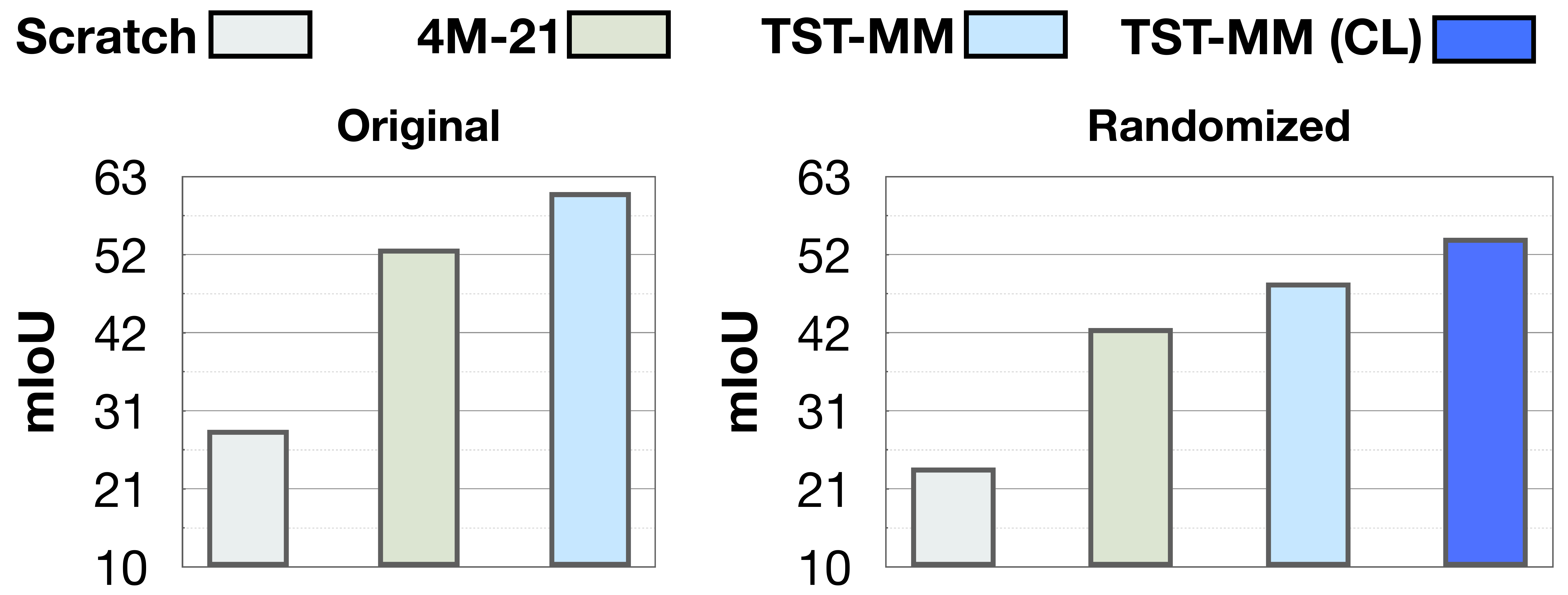}
\caption{
\textbf{Continual Learning with TST.} We study the performance of \tspmm, as the test space, undergoes lighting and minor object placement changes. The plot on the left, shows the result of the baselines on the original test space, without any changes. On the right, we present results after the test space has undergone lighting and object displacements. As expected, the \tspmm trained in the original test space, looses some performance, however as we continually train by collecting pre-training in the perturbed test space, we find that \tspmm (CL) quickly recovers performance. 
}
\label{fig:thor_cl}
\end{figure*}

\section{Continual learning with \tsp.}
\label{appendix:continual_learning}
As discussed before, when pre-training is performed on the exact same test space we deploy on, \tsp results in the most performant models. However, \tsp specializes in the test space, and all its characteristics, at the state when the pre-training data was collected. Therefore, a natural question to ask is, what happens if the test space undergoes some changes after data collection?  This could include changes in the lighting of the space or minor object placements. We begin by investigating if these changes lead to a drop in performance for the \tsp model trained on the original test space. Thereafter,  we leverage the ability of ProcTHOR to randomize object placements and lighting to create a perturbed version of the test space. Note that the overall layout and assets remain exactly the same, only the lighting and placement of small objects are varied. 

We first evaluate the performance of \tspmm pre-trained on the unperturbed test space, on the perturbed test space (Fig.~\ref{fig:thor_cl}, right), and we find that it experiences a drop as compared to its performance in the original test space, (Fig.~\ref{fig:thor_cl}, left). However, as we continually pre-train the model by collecting data in the updated test space (\tspmm (CL)), it quickly recovers the loss in performance, and is still highly performant as compared to Internet-based generalists~\citep{4m21}. This suggests that even under the condition that the test space undergoes changes, by simply continuing data collection in the test space, \tsp can continually improve its performance, without any access to external data.

\section{Test time training with \tsp}
\label{appendix:ttt-with-tsp}
\begin{wraptable}{r}{0.48\linewidth}
\vspace{-1.2em}
\centering
\begin{tabular}{@{}lcc@{}}
\toprule
MAE~\citep{He2021MaskedAA} & Before TTT & After TTT  \\  
pre-training & (mIoU $\uparrow$)  & (mIoU $\uparrow$)  \\ \midrule
Internet       & 34.54          & 39.28      \\
\tsp & \textbf{35.48}          & \textbf{42.41}      \\ \bottomrule
\end{tabular}
\caption{\textbf{\tsp with Test-Time Training.} Before TTT corresponds to the performance of the models directly after the transfer training without any test-time training, whereas after TTT shows the results when test-time training on the test samples is performed. Both models use a ViT-B backbone and are evaluated on semantic segmentation on ProcTHOR~\citep{procthor}.}
\label{tab:ttt}
\vspace{-1.0em}
\end{wraptable}
As noted in Sec.~\ref{sec:rel_work}, we share a similar goal with Test-time Training (TTT)~\citep{sun19ttt} in bridging the train-test divide. TTT does it from the lens of inference time optimization to specialize to a particular test instance, whereas \tsp attempts to specialize to a given test space by pre-training in it. 

However, in practice, these strategies can be orthogonal and complement each other. 
We can potentially apply TTT to a model pre-trained with \tsp, to improve its performance. To benchmark how this combination works, we conduct an analysis where we apply Test-time training with masked autoencoders (TTT-MAE)~\citep{gandelsman2022test}, with two pre-trained methods, MAE~\citep{He2021MaskedAA} pre-trained on Internet data and \tsprgb pre-trained on the test space. 

In TTT-MAE, we first start with a pre-trained MAE ViT-B encoder as the backbone, and train a task-specific head on the transfer set. During the test phase, the backbone is further tuned using the masked modeling objective for each test sample individually. This adaptive tuning enhances the model’s performance on the downstream task for the given test samples.

As presented in Tab.~\ref{tab:ttt}, TTT-MAE improves the mIoU results for both the Internet pre-trained backbone and \tsprgb. 
However, we find that \tsprgb gets significantly more improvement than Internet-based MAE~\citep{He2021MaskedAA}. 
Both models use a ViT-B backbone and are tested on semantic segmentation on the ProcTHOR~\citep{procthor} dataset.

\section{Sampling ratio between test space and transfer data during \method pre-training}
\label{appendix:transfer-test-ratio}
\begin{wraptable}{r}{0.58\linewidth}
\vspace{-1.1em}
\centering
\resizebox{\linewidth}{!}{%
\begin{tabular}{@{}lccccc@{}}
\toprule
\multirow{2}{*}{\begin{tabular}[c]{@{}l@{}}Model \\ Size\end{tabular}} & \multicolumn{4}{c}{\begin{tabular}[c]{@{}c@{}}Transfer set / Test space set\\ sampling ratio\end{tabular}} & \multirow{2}{*}{\begin{tabular}[c]{@{}c@{}}4M-21\end{tabular}} \\ \cmidrule(l){2-5}
                                                                       & 1/1                       & 1/4                      & 1/8                      & 1/16                     &                                                                \\ \midrule
Small                                                                  & \textbf{61.01}           & \underline{59.03}       & 56.96                    & 57.01                    & 46.12                                                         \\
Base                                                                   & 60.36                    & \underline{60.65}       & \textbf{60.85}           & 60.36                    & 53.24                                                         \\ \bottomrule
\end{tabular}%
}
\caption{\textbf{The effect of the sampling ratio between the test space and transfer data during pre-training.} We report transfer performance on semantic segmentation as we vary the sampling ratio between transfer and test space data during pre-training. There is no significant difference across ratios for the base model, while for the small model the best result is obtained with a one-to-one sampling ratio. Across all ratios, \tspmm outperforms Internet-based 4M-21 pre-training~\citep{4m21}.}
\label{tab:transfer-test-ratio}
\vspace{-0.8em}
\end{wraptable}
As mentioned in Section~\ref{sec:setup}, we found mixing RGB images from the transfer set into our pre-training data beneficial for performance. 
To study the interplay of this dataset mix further, we analyze the effect of the sampling frequency of the samples from the transfer set and the test space data during pre-training. A ratio of 1/1 implies that half the samples in pre-training come from the test space data and the other half from the transfer dataset. 
We pre-train the \tspmm model using both small and base sizes on the same test space as in Tab.~\ref{tab:max-mod-quant}, in the ProcTHOR~\citep{procthor} dataset under different ratios. The models are then transferred and evaluated on the semantic segmentation task. Tab.~\ref{tab:transfer-test-ratio} demonstrates the results for various ratio configurations and their effect on different model sizes. 
First, we find that in all cases, \tspmm consistently outperforms Internet-based 4M-21~\citep{4m21} models of the same size.
Secondly, we note that the performance of the bigger ViT-B based models is not sensitive to the ratio of sampling transfer and test space data, whereas for smaller ViT-S based models, a ratio of 1/1 seems to be a reasonable default choice. 

\section{Privacy aspect of using test-space data.}
\label{appendix:privacy_tst}
\tsp presents a risk and an opportunity, in terms of data privacy, and we believe the opportunity could be more consequential. The risk stems from the data collected in the test space, which may need to be sent out for pre-training, as on-device compute may not be sufficient. While true, this does not seem insurmountable, as \tsp can ultimately enable training a compact, dedicated model for the user space, which may become possible with some in-house compute for sensitive domains. 

On the other hand, continued commitment to the dominant paradigm of generalist models, trained on massive data, requires harvesting and pooling the data across a large number of users to create training datasets. This poses significant privacy risk — just not for the specific user that deploys the model, but for all the other users whose data had to be harvested to enable the creation of a generalist.

Taking further steps toward developing an effective test-space training paradigm with reduced need for external data, can decrease the necessity to pool diverse users’ data to make large training datasets, thereby reducing the potential privacy abuse as well as unwanted commodification of data by corporate entities. 

Besides the ethical aspect, it is interesting to more concretely understand the role of data in developing performant models (how much data, what kind of data, for pre-training or distillation, are all data equal, etc). The common assumption that large, diverse data in raw form (images and annotations) is the way to go does not have to be the full story, as several papers, including Attention Transfer~\citep{Li2024OnTS} and \tsp allude to.

\section{Experimental Setup details}
\label{appendix:implementation-details}

\subsection{Pre-training details}
\noindent \textbf{{Initialization.}}
For \tspmm, we use two initializations for pre-training. Unless stated otherwise, we pre-train our model from scratch, following the hyperparameters in Tab.~\ref{tab:supp_pretraining-setting}. Additionally, for adaptation results in Tab.~\ref{tab:max-mod-quant}, we start from a pre-trained 4M-21~\citep{4m21} model and finetune it with the hyperparameters in Tab.~\ref{tab:supp_adaptation-setting}.

\noindent \textbf{DINO Pre-training.}
For the DINO \tsp pre-training in Sec.~\ref{appendix:other-ssl}, we use the implementation from the original DINOv2 repository\footnote{\url{https://github.com/facebookresearch/dinov2}}.
We use the default provided training configuration files and train a model with the ViT-B/14 backbone for 300,000 steps with a batch size of 1024.

\begin{table}[htpb]
  \centering
    \begin{adjustbox}{max width=\linewidth}
    \begin{tabular}{@{}l|>{\centering\arraybackslash}p{6em}>{\centering\arraybackslash}p{6em}@{}}
    \toprule
    Configuration       & Small & Base              \\ 
    \midrule
    Training length ($n$ tokens) & 100B & 500B \\ 
    Warmup length ($n$ tokens)          & \multicolumn{2}{c}{10B}  \\
    Optimizer              & \multicolumn{2}{c}{AdamW \citep{loshchilov2017adamw}}                 \\
    Opt. momentum            & \multicolumn{2}{c}{$\beta_1,\beta_2=0.9,0.95$}          \\
    Base learning rate \citep{goyal2017blr}     &  \multicolumn{2}{c}{1e-4}                   \\
    Batch size             & \multicolumn{2}{c}{4096}                   \\

    Weight decay           & \multicolumn{2}{c}{0.05}                 \\
    Learning rate schedule & \multicolumn{2}{c}{Cosine decay}           \\
    Feedforward activation & \multicolumn{2}{c}{SwiGLU~\citep{shazeer2020glu}} \\
    \midrule
    Input token budget & 128 & 256 \\
    Target token budget &128 & 256 \\
    Input and target $\alpha$ & \multicolumn{2}{c}{Mixture~\citep{4m21}} \\
    Masking strategy & \multicolumn{2}{c}{Mixture~\citep{4m21}} \\
    \midrule
    Image resolution                            & \multicolumn{2}{c}{$224^2$}             \\
    Augmentation                              & \multicolumn{2}{c}{Random Crop}      \\
    Repeated sampling~\citep{Feichtenhofer2022MaskedAA} & \multicolumn{2}{c}{4} \\
    Data type & \multicolumn{2}{c}{\texttt{bfloat16}~\citep{burgess2019bfloat16}} \\

    \bottomrule
    \end{tabular}
    \end{adjustbox}   
    \caption{\textbf{Pre-training settings for scratch initialization.} Training configuration for \tspmm initialized from scratch.}
\label{tab:supp_pretraining-setting}
\end{table}


\begin{table}[htpb]
  \centering
    \begin{adjustbox}{max width=\columnwidth}
    \begin{tabular}{@{}l|>{\centering\arraybackslash}p{6em}>{\centering\arraybackslash}p{6em}@{}}
    \toprule
    Configuration       & Small & Base              \\ 
    \midrule
    Training length ($n$ tokens) & \multicolumn{2}{c}{100B} \\ 
    Warmup length ($n$ tokens)          & \multicolumn{2}{c}{10B}  \\
    Optimizer              & \multicolumn{2}{c}{AdamW \citep{loshchilov2017adamw}}                 \\
    Opt. momentum            & \multicolumn{2}{c}{$\beta_1,\beta_2=0.9,0.95$}          \\
    Base learning rate \citep{goyal2017blr}     &  \multicolumn{2}{c}{5e-5}                   \\
    Batch size             & \multicolumn{2}{c}{4096}                   \\

    Weight decay           & \multicolumn{2}{c}{0.05}                 \\
    Learning rate schedule & \multicolumn{2}{c}{Cosine decay}           \\
    Feedforward activation & \multicolumn{2}{c}{SwiGLU~\citep{shazeer2020glu}} \\
    \midrule
    Input token budget & 128 & 256 \\
    Target token budget &128 & 256 \\
    Input and target $\alpha$ & \multicolumn{2}{c}{Mixture~\citep{4m21}} \\
    Masking strategy & \multicolumn{2}{c}{Mixture~\citep{4m21}} \\
    \midrule
    Image resolution                            & \multicolumn{2}{c}{$224^2$}             \\
    Augmentation                              & \multicolumn{2}{c}{Random Crop}      \\
    Repeated sampling~\citep{Feichtenhofer2022MaskedAA} & \multicolumn{2}{c}{4} \\
    Data type & \multicolumn{2}{c}{\texttt{bfloat16}~\citep{burgess2019bfloat16}} \\

    \bottomrule
    \end{tabular}
    \end{adjustbox} 
    \caption{\textbf{Pre-training settings for Internet initialization.} Pre-training configuration for \method starting from the the pre-trained 4M~\citep{4m21} model weights.}
\label{tab:supp_adaptation-setting}
\end{table}

\subsection{Transfer details}
\label{appendix:transfer_and_eval}
\noindent \textbf{{Semantic segmentation:}}
For semantic segmentation on ProcTHOR~\citep{procthor}, Replica~\citep{straub2019replica} and Scannet++~\citep{yeshwanthliu2023scannetpp} datasets, we use the ViT encoder from the pre-trained models with a decoder head, based on the ConvNext~\citep{liu2022convnext} network with a depth of 4. This decoder head is initialized from scratch. Training details are provided in Tab.~\ref{tab:semseg-setting}. On Replica~\citep{straub2019replica}, and ProcTHOR~\citep{procthor}, pre-trained models are transferred and evaluated using a transfer training dataset of 20,000 images and evaluated on 5000 images sampled from the test space. On Scannet++~\citep{yeshwanthliu2023scannetpp}, we use a transfer dataset of 40,000 images and evaluated on 3000 images from the test space. 

\begin{table}[htpb]
\centering
  \vspace{0.5em}
\begin{tabular}{l|>{\centering\arraybackslash}p{5em}>{\centering\arraybackslash}p{5em}}
\toprule
Configuration & Small & Base \\
\midrule
Fine-tuning epochs & \multicolumn{2}{c}{64} \\
Warmup epochs & \multicolumn{2}{c}{1} \\
Optimizer &  \multicolumn{2}{c}{AdamW \citep{loshchilov2017adamw}} \\
Opt. momentum & \multicolumn{2}{c}{$\beta_1,\beta_2=0.9,0.999$}  \\
Learning rate  & 1e-4 & 2e-4 \\
Batch size & \multicolumn{2}{c}{32 (16 for Scannet++)} \\
Weight decay & \multicolumn{2}{c}{0.05}  \\
Learning rate schedule & \multicolumn{2}{c}{Cosine decay}  \\
Layer-wise lr decay~\citep{clark2020electra} & \multicolumn{2}{c}{0.75} \\
Drop path~\citep{huang2016stochdepth} & \multicolumn{2}{c}{0.1} \\
\midrule
Input resolution & \multicolumn{2}{c}{$224^2$}\\
Augmentation & \multicolumn{2}{c}{RandomFlip + RandomCrop}  \\
\bottomrule
\end{tabular}
\caption{\textbf{Semantic segmentation settings.} Configuration used for fine-tuning the pre-trained models on the semantic segmentation task.}
\label{tab:semseg-setting}
\end{table}

\noindent \textbf{{Object detection.}}
For object detection, we evaluate pre-trained models by using the ViT-based pre-trained encoder as the feature extractor in the detection framework. We use Cascade Mask-RCNN~\citep{He2017MaskRCNN,Cai2017CascadeRCNN} as our primary object detection model. Besides the feature extractor, the other learnable components including the detector's neck and head are initialized from scratch. All training and evaluations are performed using the Detectron2~\citep{wu2019detectron2} framework. Exact training settings are provided in Tab.~\ref{tab:detection-transfer}. We evaluate object detection in the test spaces from the ProcTHOR~\citep{procthor} dataset as described in Section~\ref{sec:setup}. For transfer, we use a dataset of 20,000 images from an external space, that is different from the test space. We evaluate the transferred model on 5000 images from the test space.

\begin{table}[htpb]
\centering
  \vspace{0.5em}
\begin{tabular}{l|>{\centering\arraybackslash}p{5em}>{\centering\arraybackslash}p{5em}}
\toprule
Configuration & Small & Base \\
\midrule
Fine-tuning epochs & \multicolumn{2}{c}{150} \\
Optimizer &  \multicolumn{2}{c}{AdamW \citep{loshchilov2017adamw}} \\
Opt. momentum & \multicolumn{2}{c}{$\beta_1,\beta_2=0.9,0.999$} \\
Weight decay & \multicolumn{2}{c}{0.1} \\
Learning rate  & \multicolumn{2}{c}{0.0001}\\
Learning rate schedule & \multicolumn{2}{c}{Multi-step decay} \\
Lr schedule milestones& \multicolumn{2}{c}{[Epoch 133, Epoch 144]} \\
Lr schedule decay values& \multicolumn{2}{c}{[1.0, 0.1, 0.01]} \\
Warmup epochs & \multicolumn{2}{c}{0.01} \\
Batch size & \multicolumn{2}{c}{128} \\
Layer-wise lr decay~\citep{clark2020electra} & \multicolumn{2}{c}{0.7} \\
Drop path~\citep{huang2016stochdepth} & \multicolumn{2}{c}{0.1} \\
\midrule
Input resolution & \multicolumn{2}{c}{$224^2$}\\
Augmentation & \multicolumn{2}{c}{RandomFlip + RandomCrop}  \\
\bottomrule
\end{tabular}
\caption{\textbf{Object detection settings.} Configuration used for fine-tuning the pre-trained models on the object detection task.}
\label{tab:detection-transfer}
\end{table}

\noindent \textbf{{Image Captioning.}}
For image captioning, we evaluate the pre-trained models obtained from various methods including \method, 4M-21~\citep{4m21} (Internet), and also include randomly-initialized baselines (training from scratch). We adopt a standard transformer-based encoder-decoder architecture for image captioning and employ cross-entropy loss for next-token prediction during training. Images are input to the encoder which serves as the context for the decoder network. The encoder network is initialized from the respective method's encoder while the decoder is initialized randomly. Training and hyperparameter details are listed in Tab.~\ref{tab:caption-transfer}. We also train a LLaVA style \citep{liu2023visual} model that serves as a Large-language-model-based baseline. We first train the connector module (MLP layer) using LLaVA's first-stage pretraining data consisting of 558K image-text pairs subset of the LAION-CC-SBU dataset \citep{schuhmann2021laion}. For second-stage, we re-format our ProcTHOR captioning dataset into instruction-tuning format and jointly finetune both the connector and LLM. 

\begin{figure*}
    \centering
    \includegraphics[width=0.9\textwidth] {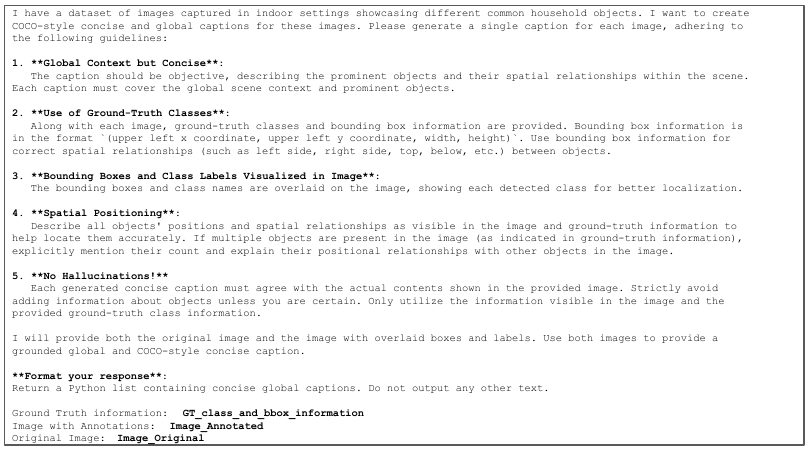}
    \caption{\textbf{LLM Prompt instruction for ProcTHOR caption generation transfer task}. We generate ground-truth captions by providing multimodal information to GPT-4o \citep{openai2023gpt4} including annotated image, class and instance-wise bounding-box information. For each image, we generate 5 COCO-style captions.}
    \label{fig:prompt_message}
\end{figure*}

\noindent \textbf{Captioning data generation.}
To train models on the captioning task, we create a transfer dataset on a set of external spaces by generating captions using GPT-4o~\citep{openai2023gpt4} for the transfer dataset. We follow a similar procedure for the evaluation set from the test space. 
We ensure the quality of generated captions by providing GPT-4o with multimodal inputs that include i) original RGB image ii) RGB image with instance-wise detection boxes and class names overlayed iii) Class names and bounding box coordinates in text format. We design an input prompt that instructs GPT-4o to leverage the multimodal inputs and generate COCO-style \citep{Lin2014MSCOCO} 5 concise captions with global context per image. For a sanity check, we randomly sampled 500 generated samples from the transfer set and found all captions to be consistent with the visual contents present in their respective images. The prompt message used for generating captions from GPT-4o is shown in {Fig.~\ref{fig:prompt_message}}.

\section{Computational Resources.}
\label{appendix:compute_resource}

All model pre-training and adaptations were done on 64 H100 GPUs, with the base and small models taking approximately 12 hours and 7 hours to train, respectively. For the semantic segmentation transfer runs, we fine-tuned the models on 4 H100 GPUs, resulting in approximately 3 hours of training for the base model and 1.5 hours for the small model. For the detection task, we only fine-tuned the base model on 8 A100 GPUs, training for approximately 6 hours. Similar to detection, for captioning we only fine-tuned the base model training on 8 H100 GPUs for approximately 6 hours.


\begin{table}[ht]
  \centering
    \begin{adjustbox}{max width=\linewidth}
    \begin{tabular}{@{}l|>{\centering\arraybackslash}p{6em}@{}}
    \toprule
    Configuration       & ProcTHOR Captioning              \\ 
    \midrule
    Fine-tuning epochs   & \multicolumn{1}{c}{1400} \\ 
    Warmup epochs         & \multicolumn{1}{c}{600}  \\
    Optimizer              & \multicolumn{1}{c}{AdamW \citep{loshchilov2017adamw}}                 \\
    Opt. momentum            & \multicolumn{1}{c}{$\beta_1,\beta_2=0.9,0.95$}          \\
    Base learning rate \citep{goyal2017blr}     &  \multicolumn{1}{c}{1e-5}                   \\
    Batch size             & \multicolumn{1}{c}{2048}                   \\

    Weight decay           & \multicolumn{1}{c}{0.05}                 \\
    Learning rate schedule & \multicolumn{1}{c}{Cosine decay}           \\
    EMA decay  & \multicolumn{1}{c}{SwiGLU~\citep{shazeer2020glu}} \\
    Eval. freq (epochs)   & \multicolumn{1}{c} {50} \\
    \midrule 
        Input resolution    & \multicolumn{1}{c}{224} \\
            
    \bottomrule
    \end{tabular}
    \end{adjustbox}
    \caption{\textbf{{Training details: Image Captioning.}} Configuration used for transfer training for image captioning.}
\label{tab:caption-transfer}
\end{table}

\begin{figure*}
    \centering
    \includegraphics[width=\textwidth]{figures/qual_fig_v3.pdf}
    \caption{\textbf{Additional qualitative results.} As demonstrated here \method performs better compared to the other models for all tasks.}
    \label{fig:quals-supmat}
\end{figure*}

\begin{figure*}
    \centering
    \includegraphics[width=\textwidth]{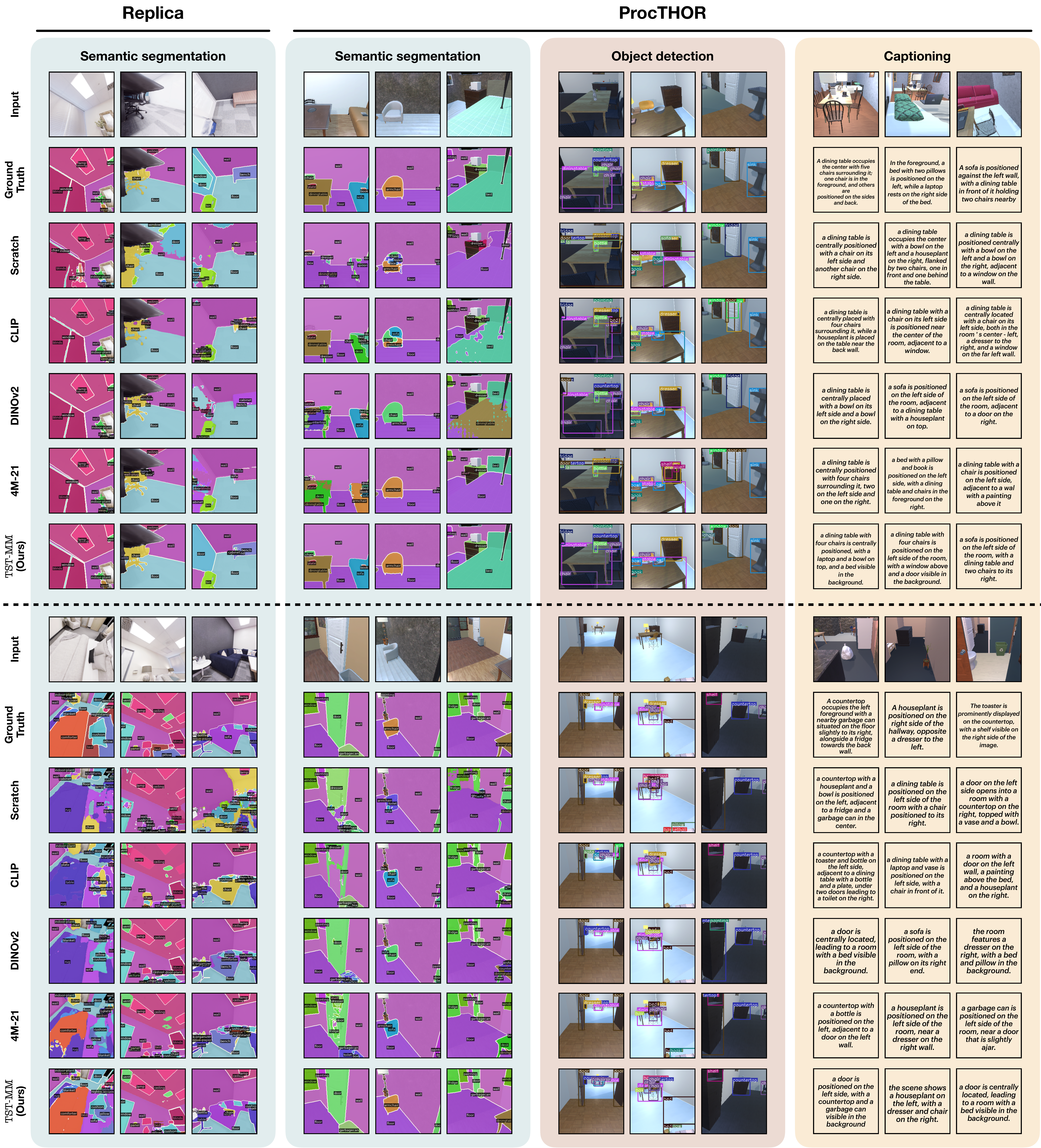}
    \caption{\textbf{Additional qualitative results.} As demonstrated here \method performs better compared to the other models for all tasks.}
    \label{fig:quals-supmat-2}
\end{figure*}

\clearpage